\documentclass[lettersize,journal]{IEEEtran}
\usepackage{amsmath,amsfonts}
\usepackage{algorithmic}
\usepackage{array}
\usepackage[caption=false,font=normalsize,labelfont=sf,textfont=sf]{subfig}
\usepackage{textcomp}
\usepackage{stfloats}
\usepackage{url}
\usepackage{verbatim}
\usepackage{graphicx}
\def\BibTeX{{\rm B\kern-.05em{\sc i\kern-.025em b}\kern-.08em
    T\kern-.1667em\lower.7ex\hbox{E}\kern-.125emX}}
\usepackage{balance}

\usepackage{lineno}
\usepackage{algorithm}
\usepackage{booktabs}
\usepackage{float}
\usepackage{multicol}
\usepackage{rotating}
\usepackage{blindtext}
\usepackage{multirow}
\usepackage[dvipsnames]{xcolor}

\usepackage{makecell}
\usepackage{setspace}
\usepackage{arydshln}
\usepackage{hyperref}
\usepackage{ulem}  
\usepackage{pgfplots}
\pgfplotsset{compat=1.17}
\usepackage{pgfplotstable}

\begin{document}
\title{\textit{DiverseNet}: Decision Diversified Semi-supervised Semantic Segmentation Networks for Remote Sensing Imagery}

\author{
Wanli Ma\thanks{}, 
Oktay Karakuş, 
Paul L. Rosin \\
School of Computer Science and Informatics, Cardiff University, Cardiff, U.K. 
}

\markboth{Journal of \LaTeX\ Class Files,~Vol.~18, No.~9, September~2020}%
{How to Use the IEEEtran \LaTeX \ Templates}

\maketitle

\begin{abstract}
Semi-supervised learning (SSL) aims to help reduce the cost of the manual labelling process by leveraging a substantial pool of unlabelled data alongside a limited set of labelled data during the training phase. Since pixel-level manual labelling in large-scale remote sensing imagery is expensive and time-consuming, semi-supervised learning has become a widely used solution to deal with this. However, the majority of existing SSL frameworks, especially various teacher-student frameworks, are too bulky to run efficiently on a GPU with limited memory. There is still a lack of lightweight SSL frameworks and efficient perturbation methods to promote the diversity of training samples and enhance the precision of pseudo labels during training. In order to fill this gap, we proposed a simple, lightweight, and efficient SSL architecture named \textit{DiverseHead}, which promotes the utilisation of multiple decision heads instead of multiple whole networks. Another limitation of most existing SSL frameworks is the insufficient diversity of pseudo labels, as they rely on the same network architecture and fail to explore different structures for generating pseudo labels. To solve this issue, we propose \textit{DiverseModel} to explore and analyse different networks in parallel for SSL to increase the diversity of pseudo labels. The two proposed methods, namely \textit{DiverseHead} and \textit{DiverseModel}, both achieve competitive semantic segmentation performance in four widely used remote sensing imagery datasets compared to state-of-the-art semi-supervised learning methods. Meanwhile, the proposed lightweight DiverseHead architecture can be easily applied to various state-of-the-art SSL methods while further improving their performance. The code is available at https://github.com/WANLIMA-CARDIFF/DiverseNet.
\end{abstract}


\section{Introduction}

Supervised deep learning has become the dominant technique in computer vision during the last decade. Building on its success in computer vision, many remote sensing applications, such as land cover classification, change detection and object detection have seen significant improvements as similar tasks \cite{vali2020deep, zhang2024integrating, wang2025remote, zhang2024adaptive, hu2024contrastive}. Nevertheless, supervised learning necessitates a substantial and meticulously labelled dataset. In the case of extensive remote sensing data, such as satellite imagery and drone-captured images in complex terrains, acquiring pixel-wise expert annotations is a time-consuming, labour-intensive, and costly process. While the field of computer vision provides numerous well-annotated datasets, transferring deep learning models trained on these datasets to the remote sensing domain is a formidable challenge. This is mainly due to the substantial differences between typical computer vision images and remote sensing data, such as hyperspectral and synthetic aperture radar (SAR) imagery, which often exhibit unconventional and non-intuitive characteristics. To address the issue of limited access to massive labelled datasets, SSL offers a viable solution by using a small amount of labelled data while leveraging the abundance of unlabelled data \cite{wang2022semi, yang2023label, wang2023lithological}. This is because large volumes of unlabelled imagery generally can be easily and freely accessed from open-access remote sensing data sources. 

\begin{figure}
\centering
\includegraphics[width=0.5\textwidth]{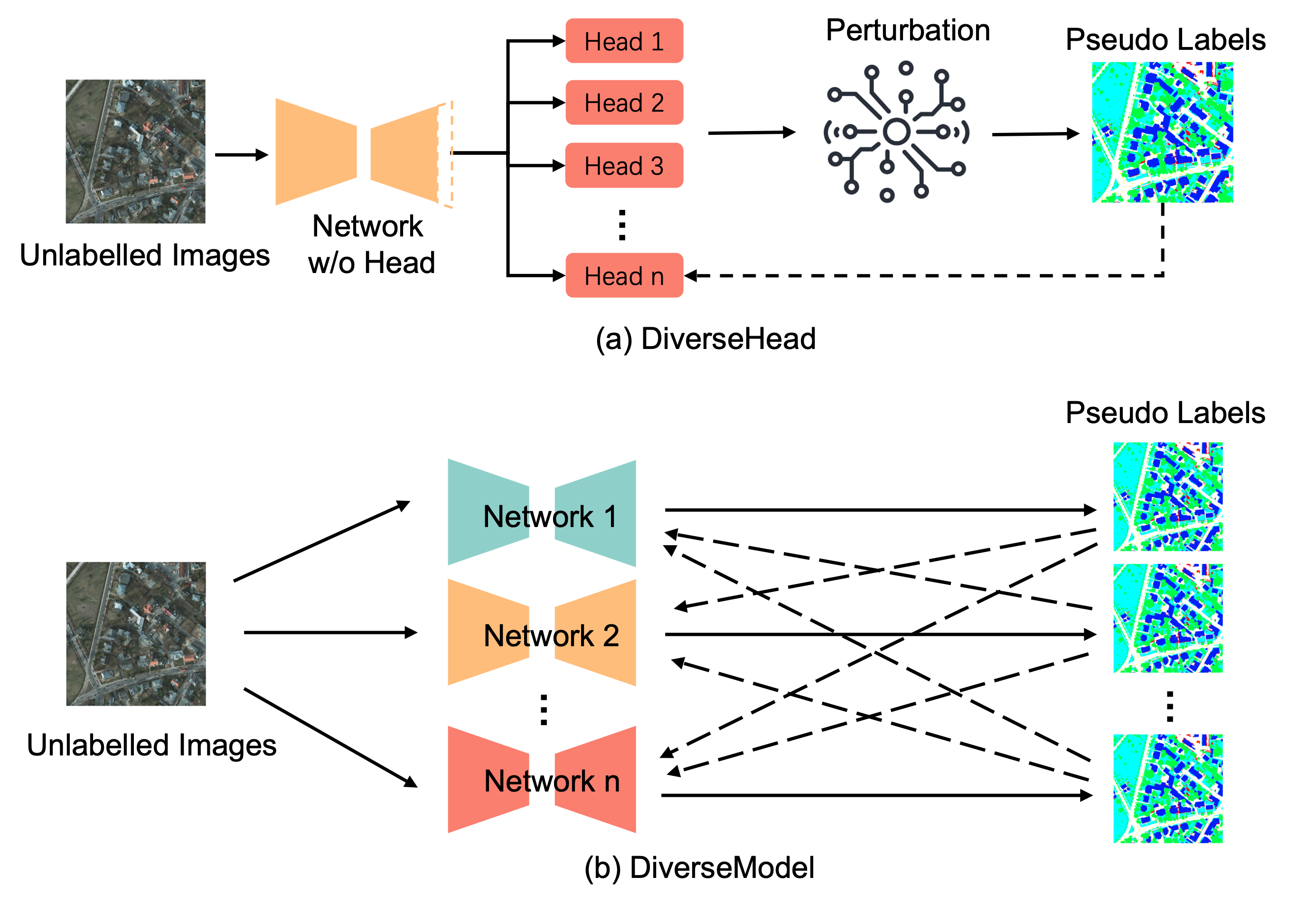}
\caption{Two kinds of pseudo label generation and usage methods for SSL based on (a) DiverseHead with multiple heads and (b) DiverseModel with multiple models. `$\longrightarrow$‘ means data stream, `$\dashrightarrow$‘ means loss supervision. The \textit{`dynamic freezing}' and \textit{`dropout'} are used as perturbation methods in the DiverseHead framework. }
\vspace{-15pt}
\label{Fig_title}
\end{figure}

SSL has become a widely used technique in computer vision to reduce the labour-intensive and costly annotation process for various applications such as image classification \cite{zhang2021flexmatch, huang2021embedding, wang2021semi2} and segmentation \cite{hu2021semi, wang2022semi_u2pl, wang2021semi}, whilst leading to competitive performance. Specifically, taking advantage of ``\emph{pseudo}" labels generated by the prediction of unlabelled data has become a mainstream class of SSL methods. Thus, the quality of pseudo labels becomes a crucial factor in determining the training effectiveness of the models. With the success of SSL in computer vision, many SSL semantic segmentation approaches for remote sensing imagery have been explored, especially during the last decade \cite {wang2021ranpaste, zhang2022semi, lu2022simple}. Specifically, based on consistency learning, U-MCL \cite{lu2025uncertainty} proposes an uncertainty-aware technique integrating masks for SSL semantic segmentation of remote sensing imagery, and \cite{zhang2024bias} promotes the importance of bias-correction .

In the SSL literature, enhancing the accuracy and diversity of pseudo labels has become a major challenge and a central research focus. This is due to the fact that pseudo label accuracy is often inadequate, especially when the labelled training data is limited or incomplete. On the other hand, enhancing the diversity of pseudo labels becomes another key research focus in SSL to improve model robustness, particularly in the context of consistency regularisation. \cite{french2019semi}. 


In order to improve the quality of pseudo labels and empower networks to harness the potential of unlabelled data, a notable technique known as consistency regularisation has emerged as a widely adopted method for SSL. \cite{french2019semi, filipiak2021n}. Specifically, consistency regularisation methods are built up on the theory of assumption of smoothness, which suggests that if two points lie in a high-density region of feature space and are close to each other, their corresponding labels should be the same or consistent \cite{chen2010semi,luo2018smooth,french2019semi}. In practice, consistency regularisation SSL executes this assumption by forcing networks to produce consistent predictions for modified versions of unlabelled data or their features using various perturbation techniques.

The strategies for the aforementioned perturbations can be categorised into three groups, namely \textit{input, feature,} and \textit{network} perturbations. Input perturbation involves modifying or altering input data, enabling SSL approaches to enforce consistency in predictions for these altered inputs. The widely used input perturbation method is adding artificial noise to input images. 
However, it might lead to incorrect or noisy pseudo labels for unlabelled examples and negatively impact the training efficiency by providing incorrect guidance to the model when training with these pseudo labels. Apart from input perturbation, feature perturbation methods introduce noise to both low- and high-level features. These perturbed features are then fed into multiple decoders to generate multiple outputs, followed by the enforcement of consistency among the outputs obtained from different decoders. However, similar to input perturbation, feature perturbation potentially generates inaccurate representations due to introducing noise that fails to accurately capture the underlying patterns in the original data. This may result in incorrect or noisy pseudo-labels for unlabelled examples, ultimately hindering the learning process. Lastly, network perturbation uses multiple networks to promote diversity of predictions \cite{chen2021semi}. Unlike the previous two perturbation methods, network perturbation techniques introduce perturbations in a more structured manner, generated by the model itself instead of artificial noise. However, they generally require significantly more computational resources due to the greater number of complete networks (or their internal stages) involved. 

To further justify our design, our network perturbation strategies (dynamic freezing and dropout) introduce structured, model-driven diversity without adding external noise. Unlike input and feature perturbations, which risk corrupting pseudo-labels or internal representations due to artificial noise injection, our approach preserves semantic integrity while enhancing prediction diversity. Furthermore, DiverseHead applies lightweight perturbations to the model head, achieving a balance between diversity and efficiency without the heavy computational cost of maintaining multiple full models.

The previous overview emphasises that while earlier perturbation techniques in SSL provide certain benefits, they also have inherent limitations, including inefficiencies in generating high-quality pseudo labels and a high computational cost. Facing the challenges posed by perturbation-based SSL and its complexity, it becomes imperative to explore more efficient and lightweight approaches. This work proposes two perturbation-based semi-supervised network architectures, coined as \textit{DiverseNet}, which consist of multiple head (DiverseHead) and multiple model (DiverseModel) based SSL frameworks for various semantic segmentation applications of remote sensing imagery. A brief demonstration of the proposed lightweight SSL framework called \textit{DiverseHead} is shown in Figure \ref{Fig_title}-(a). 
In addition, we also further analysed a previously proposed cross-network based SSL structure called \textit{DiverseModel} \cite{ma2023confidence}, as shown in Figure \ref{Fig_title}-(b) for scenarios equipped with high-memory computational resources. Specifically, the contributions of this work are as follows:
\begin{enumerate}
    \item We introduce DiverseHead, a simple, lightweight, and efficient SSL framework which employs multiple decision heads within a single network. This structure is inspired by bagging (also called bootstrap aggregating), which helps enhance pseudo label quality by integrating perturbed parameters and features within the network architecture. 

    \item  To introduce perturbation for diversifying decisions, we incorporate two key techniques, \textit{dynamic freezing} and \textit{dropout}, into the DiverseHead architecture, aiming to diversify the network's parameters and high-level features, respectively. The proposed perturbation strategies, incorporating multiple heads, are readily applicable to a variety of state-of-the-art SSL methods and can further enhance their performance.
 
    \item We propose a dual voting mechanism, Mean Voting and Max Voting, to aggregate multihead predictions and produce high-fidelity pseudo labels for DiverseHead. This mechanism leverages both collective consensus and individual confidence to further enhance pseudo-label robustness during training.

    \item  We provide a more detailed comparison study for a previously proposed architecture DiverseModel \cite{ma2023confidence} on various semantic segmentation datasets in this paper. Also, we use Grad-CAM \cite{selvaraju2017grad} to verify the observation that different networks exhibit varied attention to the same input. 
\end{enumerate}

The rest of the paper is organised as follows: Section \ref{subsec:related} discusses the related work on semi-supervised semantic segmentation in remote sensing and some basic knowledge on ensemble machine learning whilst in Section \ref{sec:proposed1} and \ref{sec:proposed2}, the proposed algorithms DiverseHead and DiverseModel are presented.  Section \ref{sec:Datasets} describes the utilised segmentation dataset of remote sensing imagery. The experimental setting along with both the qualitative and quantitative analyses of the results, are presented in Section \ref{sec:results}. Section \ref{sec:conc} concludes the paper with a summary.

\section{Related Work}\label{subsec:related}
\textbf{Semantic segmentation} is rapidly developing in remote sensing with the success of deep learning in computer vision. Due to the strong task similarity, semantic segmentation techniques are used for various remote sensing applications, such as land cover classification/mapping \cite{dong2019land}, building change detection \cite{zheng2022hfa}, road extraction \cite{ghandorh2022semantic}, and marine debris detection \cite{kikaki2022marida, booth2023high}. Specifically, Fully Convolutional Networks (FCNs) \cite{long2015fully} have made a considerable contribution to various segmentation tasks either in remote sensing or computer vision. Following the FCNs' success, SegNet \cite{badrinarayanan2017segnet} and UNet \cite{ronneberger2015u} adopt a symmetrical encoder-decoder structure with skip connections, leveraging multi-stage features within the encoder. Alternatively, PSPNet \cite{zhao2017pyramid} introduces a pyramid pooling structure that helps provide a global contextual understanding for pixel-level scene parsing. The DeepLab architecture \cite{chen2017deeplab} introduces atrous convolution and atrous spatial pyramid pooling (ASPP), allowing the network to adjust the spatial receptive field of convolution kernels by using different dilation rates. Then, DeepLab was extended to DeepLabv3+ \cite{chen2018encoder} with an improved encoder-decoder structure, which is helpful to refine segmentation results, especially around object boundaries~\cite{xia2021cloud, zhang2019multi}. Recently, GLOTS \cite{liu2023rethinking} was proposed for semantic segmentation of remote sensing images, aiming to acquire consistent feature representations by leveraging transformers in both the encoder and decoder. DeepLabv3+ is one of the most widely used networks in the literature for semi-supervised learning segmentation in the computer vision area.

\textbf{Semi-supervised learning} aims to alleviate the need for expensive annotation work by making use of both labelled and unlabelled data. Self-training \cite{lee2013pseudo} (also known as pseudo labelling) represents one of the primitive SSL strategies for both classifications \cite{kim2020distribution} and segmentation \cite{zhu2021improving}. It generates pseudo-labels using model predictions for unlabelled data, which are then utilised to retrain the model. Another widely developed SSL approach called consistency regularisation \cite{french2019semi} is to force networks to give consistent predictions for unlabelled inputs that undergo diverse perturbations. In the context of the remote sensing field, Lu et al. \cite{lu2023weak} propose a weak-to-strong consistency learning for semi-supervised semantic segmentation. Building on the weak-to-strong consistency learning, Lv et al. \cite{lv2024advancing} further explore the efficient exploitation of labelled and
unlabelled images. Additionally, MIMSeg \cite{cai2024consistency} integrates weak-to-strong consistency learning with masked image modelling, and DWL \cite{huang2024decouple} combines it with a decoupled weighting learning framework for semi-supervised semantic segmentation of remote sensing imagery.

In computer vision, CCT \cite{ouali2020semi} employs an encoder-decoder architecture with multiple auxiliary decoders. These decoders introduce diversity in the output by feature perturbations specific to each auxiliary decoder. They calculate the MSE loss between the predictions of the main decoder and each auxiliary decoder without creating pseudo labels. It is worth noting that the unsupervised loss is not used to supervise the main decoder. 
Following CCT, subsequent methods like GCT \cite{ke2020guided} and CPS \cite{chen2021semi} have been proposed to introduce network perturbation for consistency regularisation. Both of them use the same network structures but with different weight initialisation. CPS differs from GCT by using pseudo-labels generated from two networks to enforce consistency, whereas GCT achieves consistency regularisation by minimising the loss between the probability predictions of networks working in conjunction with a flow detector. Although the network perturbation with different weight initialisation provides some diversity for pseudo labels in GCT and CPS, the ability to generate diversity is still limited. Another group of SSL methods is built upon the teacher-student architecture. As a classic teacher-student model, MT \cite{tarvainen2017mean} uses an EMA of the student's weights for the teacher, trains student models with augmented inputs, and applies a consistency cost to align their outputs. ICNet \cite{wang2022semi} was proposed to use teacher networks to improve the quality of pseudo labels and increase the model difference based on an iterative contrast network. Building on the teacher-student architecture, some of its variations have become SOTA. For example, iMAS \cite{zhao2023instance} highlights the importance of instance differences and introduces instance-specific and model-adaptive supervision for semi-supervised semantic segmentation. AugSeg \cite{zhao2023augmentation} employs an enhanced data augmentation technique for input data and adaptively injects labelled information into unlabelled samples based on the model's estimated confidence in each sample. These methods consistently rely on the same network architecture with different weights. The previously proposed DiverseModel argues that
advocating the use of different models can obtain distinct and complementary features from these models, even with the same input data. Specifically, DiverseModel explores different networks in parallel to generate more diversity of pseudo labels to improve training effectiveness. Although these discussed methods achieve competitive performance, they consistently rely on multiple networks, making them bulky and potentially unsuitable for users with limited computational resources.

\textbf{Ensemble machine learning} is a concept that employs multiple learners and combines their predictions \cite{sewell2008ensemble}. Bagging, a form of ensemble learning, is a technique aimed at reducing prediction variance by creating multiple iterations of a predictor and then utilising them to form an aggregated predictor \cite{breiman1996bagging}. Specifically, bagging creates sample subsets by randomly selecting from the training dataset and subsequently utilises these acquired subsets to train the foundational models for integration. When predicting a numerical outcome, aggregation is done by averaging the different versions, whereas for class prediction, it is based on a majority vote. Bagging is a commonly employed approach for enhancing the robustness and precision of machine learning algorithms for classification and regression \cite{ren2016ensemble}. 

\section{DiverseHead (Cross-head Supervision)}\label{sec:proposed1}

\begin{figure*}
\centering
\includegraphics[width=0.7\textwidth]{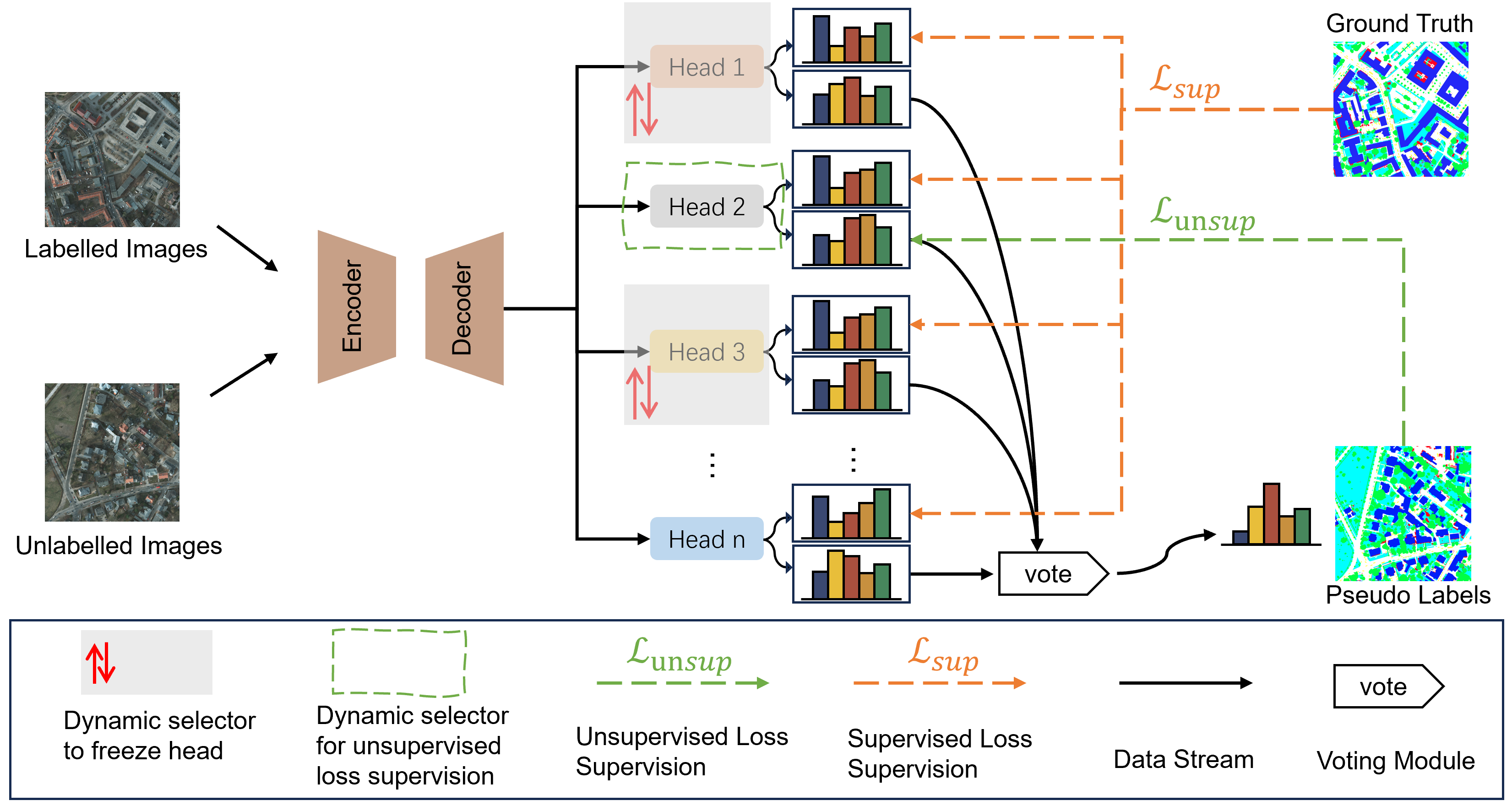}
\caption{DiverseHead: an online semi-supervised learning approach. This figure applies the dynamic freezing strategy: the freezers (Dynamic Selector in the figure) randomly select a certain number of heads to freeze the parameter of heads (not updated by backpropagation). Additionally, during every iteration, all heads undergo supervision through a supervised loss, yet each head is randomly chosen to be updated by an unsupervised loss.}
\label{fig_1}
\end{figure*}

\begin{figure}
\centering
\includegraphics[width=0.475\textwidth]{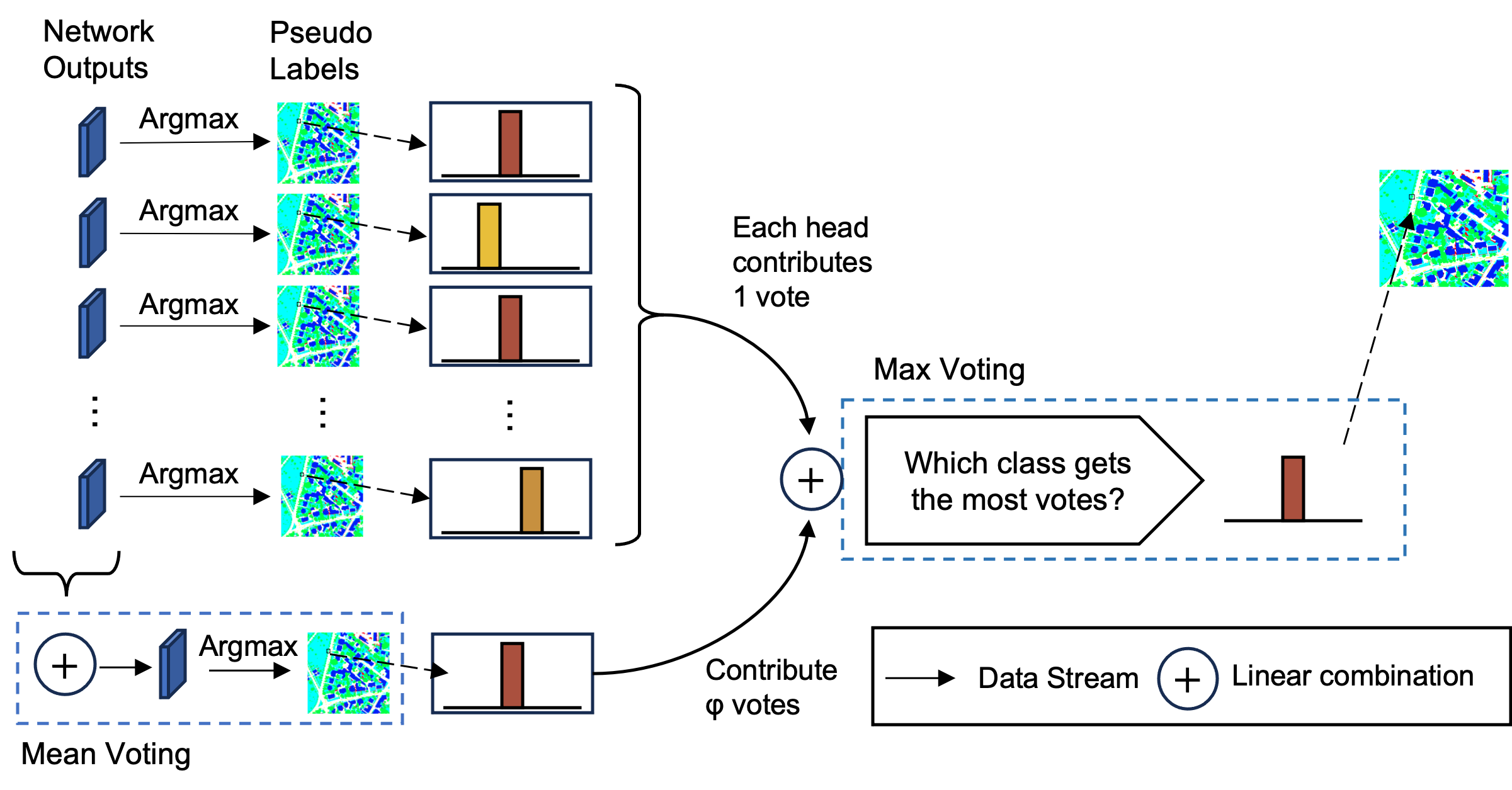}
\caption{The Proposed Voting Module: a voting mechanism for the pseudo label creation. In the unsupervised part, the voting module combines the mean output of multiple heads (mean voting) and individual pseudo labels (max voting) to generate more efficient pseudo labels. Argmax returns the indices of the maximum values of the prediction along the class dimension. The dashed arrow serves as an illustration of a pixel voting for its classification in a segmentation map. }
\vspace{-10pt}
\label{fig:Voting_Module}
\end{figure}

\begin{figure}
\centering
\includegraphics[width=0.475\textwidth]{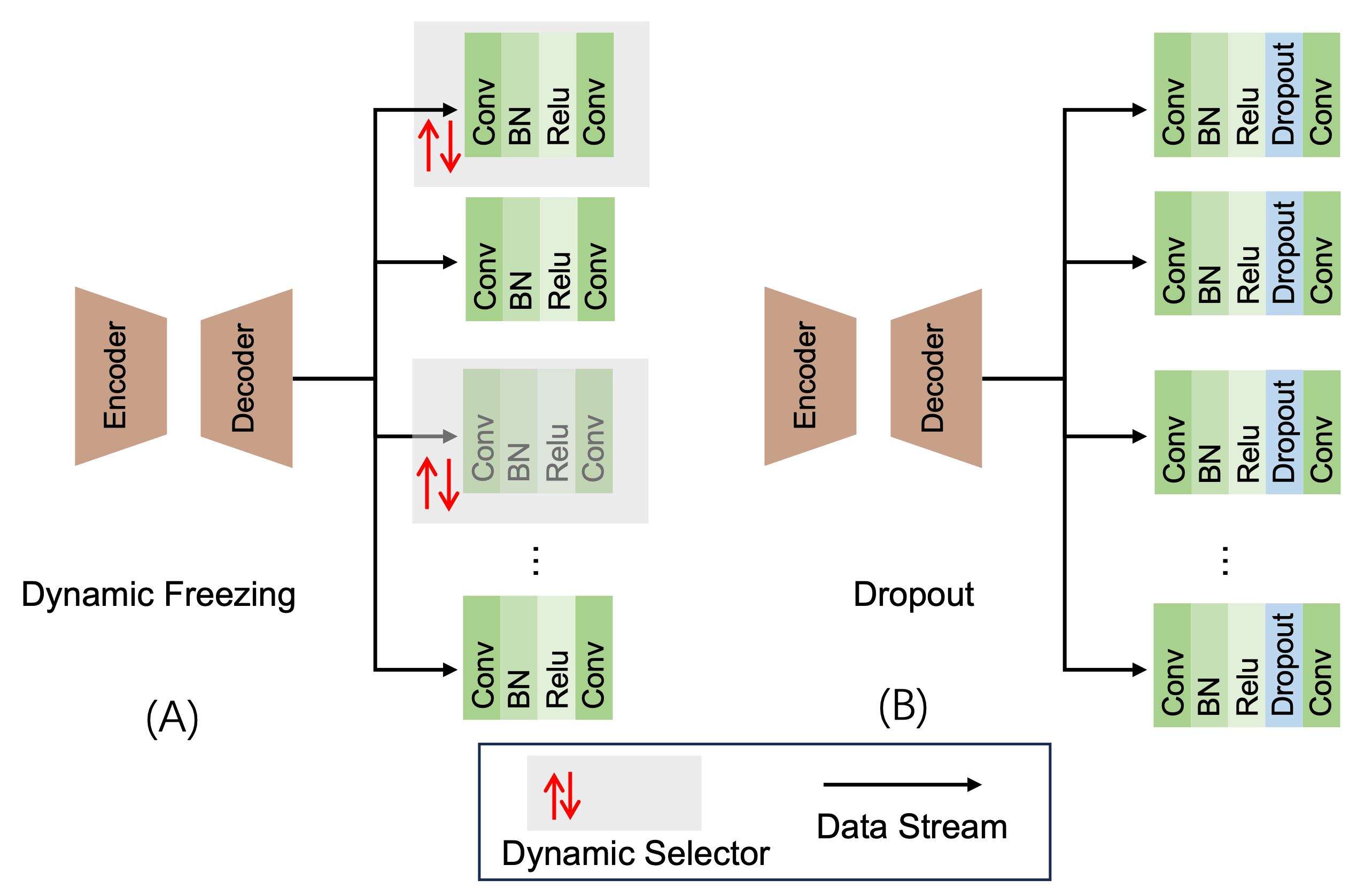}
\caption{The proposed two perturbation methods: (A) Dynamic Freezing and (B) Dropout. Dynamic Freezing was designed to enhance the parameter diversity across multiple heads. During each training iteration, a specific subset of heads is randomly selected (The Dynamic Selector in the figure is used for this purpose), and their parameters are frozen, meaning they are not updated by minimising the loss in that iteration. These parameters are unfrozen before the next iteration begins. Instead, for Dropout, each channel of features passed through each head is independently zeroed out with a dropout rate $p$ during each forward pass.}
\vspace{-5pt}


\label{fig:Two_perturbation_methods}
\end{figure}

This section introduces the details of the proposed DiverseHead SSL method, which uses a single network with multiple heads, illustrated in Figure~\ref{fig_1}. Each head has 2 convolutional layers. Unlike CCT, the proposed DiverseHead method treats all heads equally, avoiding a distinction between the main and auxiliary heads. Meanwhile, all heads benefit from labelled data by applying supervised losses to each. This helps create better pseudo labels in the subsequent steps of the proposed method. On the other hand, rather than adding artificial noise to perturb head features as in CCT, DiverseHead introduces two perturbation strategies: Dynamic Freezing and Dropout. They are demonstrated in Figure \ref{fig:Two_perturbation_methods} and explained in detail in Section \ref{subsec:perturbation_methods}. During each training iteration, in addition to supervised losses, an unsupervised loss is computed between the pseudo-label and the prediction. The pseudo label is derived from an efficient voting module illustrated in Figure \ref{fig:Voting_Module}. Although non-differentiable operations (e.g. Voting and Argmax) could potentially limit optimisation during training, in our framework, parameter perturbations, feature perturbations, and independent initialisation across heads and models inject sufficient diversity and stochasticity into the training process. These factors enable effective gradient-based optimisation despite the presence of non-differentiable components. The proposed method can be seen as a combination of self-training and consistency regularisation, leveraging model predictions to supervise itself and forcing all perturbed heads to produce consistent outputs.

To provide a detailed description of the proposed semi-supervised framework, given both a labelled data set $\mathcal{B}^l=\left\{\left(x_i, y_i\right)\right\}_{i=1}^M$ containing $M$ images and an unlabelled data set $\mathcal{B}^u=\left\{u_i\right\}_{i=1}^N$ with $N$ images, the network $Q$ is constructed with multiple heads denoted as $\left\{head^i\right\}_{i=1}^L$, where \mbox{\it L} is the number of heads. Each of these heads is initialised differently. The proposed SSL approach expects to gain the trained network $Q$ leveraging the labelled and unlabelled data.

The output of each head refers to a probability-based prediction for all classifications. When working with labelled data, the supervised loss $\mathcal{L}_{sup,s}^j$ between the $s'th$ ground truth $y_s$ and its corresponding prediction $p_s^j$ for $j'th$ head is defined by using the standard cross-entropy loss function $\ell_{ce}$:

\begin{equation}
\label{equ:equ1}
\mathcal{L}_{sup,s}^j = \frac{1}{W \times H} \sum_{i=1}^{W \times H}\ell_{c e}\left(p_{i,s}^j, y_{i,s}\right),
\end{equation}
where $W$ and $H$ refer to the width and height of input images. The final supervised loss for the $s'th$ labelled data is determined by the mean of the losses across all heads

\begin{equation}
\label{equ:equ2}
\mathcal{L}_{sup,s} = \frac{1}{L} \sum_{j=1}^{L} \mathcal{L}_{sup,s}^j
\end{equation}
where \mbox{\it L} is the number of heads.

In each iteration, we also get a prediction $r_k^j$ corresponding to the $k'th$ unlabelled data $u_k^j$ from the $j'th$ head. Inspired by ensemble learning, a voting module is proposed to get high-precision pseudo labels. There are two voting mechanisms in the proposed voting module, called mean voting and max voting, respectively shown in Figure \ref{fig:Voting_Module}. The former aggregates the predictions from all heads to create a combined prediction $\hat{r}_k^{mean}$. Subsequently, this combined prediction is used to calculate the mean pseudo labels $\hat{y}_k^{mean}$ through an $argmax$ operation, which returns the indices of the maximum values of the prediction along the class dimension. Apart from the mean voting module, the individual pseudo label is generated from the output of each head by the $argmax$ function. The latter regards all pseudo labels as voters, in which the mean pseudo label contributes $\varphi$ weight and each individual pseudo label contributes unit weight. $\varphi$ is a learnable parameter and its value changes depending on the dataset and training process. The max voting module returns the class number that gets the most votes for each pixel. After max voting, an optimal pseudo label $\hat{y}_k^{final}$ is created to calculate unsupervised loss $\mathcal{L}_{unsup,k}$ by using the cross-entropy loss function:

\begin{equation}
\label{equ:equ3}
\mathcal{L}_{unsup,k} = \frac{1}{W \times H} \sum_{i=1}^{W \times H}\ell_{c e}\left(r_{i,k}^{main}, \hat{y}_{i,k}^{final}\right),
\end{equation}
where $r_k^{main}$ is the prediction from the randomly selected single head. 

Finally, the whole loss can be written as

\begin{equation}
\label{equ:equ4}
\mathcal{L} = \mathcal{L}_{sup,s} + \lambda \mathcal{L}_{unsup,k},
\end{equation}
where $\lambda$ is the trade-off weight between the supervised and unsupervised losses. We use different subscripts ($s$ and $k$) for the supervised and unsupervised losses, respectively, to account for the difference in the numbers of labelled and unlabelled data. Specifically, the number of labelled data samples is smaller than that of unlabelled data samples. The solution to this issue is that, in each epoch, the labelled data set is repeatedly used in cycles for multiple iterations until all unlabelled data have been processed once.

\subsection{Perturbation Methods}
\label{subsec:perturbation_methods}

 \begin{algorithm}[h]
 \small
\caption{\small DiverseHead Semi-Supervised Learning with Dynamic Freezing
Pseudocode. The Labelled Training Dataset is defined as $\mathcal{B}^l$. Since the number of labelled data is smaller than that of unlabelled data, the labelled data is used in cycles for one epoch. We define it as $\textit{cycle} \left(\mathcal{B}^l\right)$.}\label{alg:alg1}
\begin{algorithmic}
\STATE 
{\textsc{\textbf{Initialization:}}} 
\STATE \hspace{1.4cm} Randomly initialise model Q
\STATE \hspace{1.4cm} Initialise backbone using ResNet-50
\STATE {\textsc{\textbf{INPUT:}}} Labelled Training Dataset $\mathcal{B}^l=\left\{\left(x_i, y_i\right)\right\}_{i=1}^M$
\STATE \hspace{1.4cm} Unlabelled Training Dataset $\mathcal{B}^u=\left\{u_i\right\}_{i=1}^N$ 
\STATE $L = length(heads)$
\STATE  \textbf{for} $ \left\{\left(x_s, y_s\right),u_k\right\}_{k=1, s=k\%M}^N \in  \left\{\textit{cycle} \left(\mathcal{B}^l\right) , \mathcal{B}^u\right\}$ \textbf{do} 
\STATE \hspace{0.5cm} $\mathcal{R} = Randint(0,L,\frac{1}{2}L)$
\STATE \hspace{0.5cm} $Freeze(\left\{head^i\right\}_{i \in \mathcal{R}})$
\STATE \hspace{0.5cm} \textbf{for} $ j \in \left\{1,...,L\right\}$ \textbf{do}
\STATE \hspace{1.0cm} $  p_s^j = Q^{head_j}(x_s^j)$
\STATE \hspace{1.0cm} $ \mathcal{L}_{sup,s}^j = loss(p_s^j, y_s) $ based on (\ref{equ:equ1})
\STATE \hspace{1.0cm} $ r_k^j = Q^{head_j}(u_k^j)$
\STATE \hspace{1.0cm} $ \hat{y}_k^{j} \gets argmax(r_k^{j})$
\STATE \hspace{0.5cm} $ r_k^{mean} = sum\left(\left\{r_k^{j}\right\}_{j=1}^{L}\right)$
\STATE \hspace{0.5cm} $ \hat{y}_k^{mean} \gets argmax(r_k^{mean})$
\STATE \hspace{0.5cm} $ \hat{y}_k^{final} \gets voting(\left\{\hat{y}_k^{j}\right\}_{j=1}^{L},\hat{y}_k^{mean})$
\STATE \hspace{0.5cm} $ r_k^{main} \gets sample\left(\left\{r_k^1, r_k^2, ..., r_k^{L}\right\}\right)$ 
\STATE \hspace{0.5cm} $ \mathcal{L}_{unsup,k} \gets loss(r_k^{main}, \hat{y}_k^{final}) $ based on (\ref{equ:equ3})
\STATE \hspace{0.5cm} $ \mathcal{L} \gets \frac{1}{L} \sum_{j=1}^{L} \mathcal{L}_{sup,s}^j + \lambda \mathcal{L}_{unsup,k} $ 
\STATE \hspace{0.5cm} Minimize $\mathcal{L}$ to update model $Q$
\STATE \hspace{0.5cm} $\mbox{\textit{Unfreeze}}(\left\{head^i\right\}_{i=R})$
\STATE {\textsc{\textbf{OUTPUT:}}} Trained model $Q$
\end{algorithmic}
\end{algorithm}

Based on the framework of DiverseHead, we propose a parameter perturbation method called dynamic freezing as shown in Figure \ref{fig:Two_perturbation_methods} (a). 
The pseudocode of the algorithm for DiverseHead with the perturbation of dynamic freezing is given in \textbf{Algorithm} \ref{alg:alg1}. Specifically, we use DeepLabv3+ with a ResNet-50 backbone as the segmentation network in our framework, ensuring consistency with the methods compared in this paper. The backbone is pretrained on ImageNet, and other parameters are initialised by Kaiming Initialization. Both labelled and unlabelled data are input to the framework, then the first step of dynamic freezing refers to a process where half of the heads are randomly selected to be frozen during each iteration. This implies that the parameters within these selected heads will remain unchanged and not undergo updates for the current iteration. Every head has an equal probability of being chosen. Following the cross-head supervision above, the segmentation model is updated by the supervised and unsupervised losses. Before going to the next iteration, those frozen heads are unfrozen. 

\begin{algorithm}
\small
\caption{\small DiverseHead Semi-Supervised Learning with Dropout Pseudocode. The Labelled Training Dataset is defined as $\mathcal{B}^l$. Since the number of labelled data is smaller than that of unlabelled data, the labelled data is used in cycles for one epoch. We define it as $\textit{cycle} \left(\mathcal{B}^l\right)$.}\label{alg:alg2}
\begin{algorithmic}
\STATE 
\STATE {\textsc{\textbf{Initialization:}}} \STATE \hspace{1.4cm} Randomly initialise model Q
\STATE \hspace{1.4cm} Initialise backbone using ResNet-50
\STATE \hspace{1.4cm} Add dropout layer before the last convolutional layer
\STATE \hspace{1.4cm} for each head
\STATE {\textsc{\textbf{INPUT:}}} Labelled Training Dataset $\mathcal{B}^l=\left\{\left(x_i, y_i\right)\right\}_{i=1}^M$
\STATE \hspace{1.4cm} Unlabelled Training Dataset $\mathcal{B}^u=\left\{u_i\right\}_{i=1}^N$ 
\STATE $L = length(heads)$
\STATE \textbf{for} $ \left\{\left(x_s, y_s\right),u_k\right\}_{k=1,s=k\%M}^N \in \left\{\textit{cycle} \left(\mathcal{B}^l\right) , \mathcal{B}^u\right\}$ \textbf{do}
\STATE \hspace{0.5cm} \textbf{for} $ j \in \left\{1,...,L\right\}$ \textbf{do}
\STATE \hspace{1.0cm} $  p_s^j = Q^{head_j}(x_s^j)$
\STATE \hspace{1.0cm} $ \mathcal{L}_{sup,s}^j = loss(p_s^j, y_s) $ based on (\ref{equ:equ1})
\STATE \hspace{1.0cm} $ r_k^j = Q^{head_j}(u_k^j)$
\STATE \hspace{1.0cm} $ \hat{y}_k^{j} \gets argmax(r_k^{j})$
\STATE \hspace{0.5cm} $ r_k^{sum} = sum\left(\left\{r_k^{j}\right\}_{j=1}^{L}\right)$
\STATE \hspace{0.5cm} $ \hat{y}_k^{mean} \gets argmax(r_k^{mean})$
\STATE \hspace{0.5cm} $ \hat{y}_k^{final} \gets voting(\left\{\hat{y}_k^{j}\right\}_{j=1}^{L},\hat{y}_k^{mean})$
\STATE \hspace{0.5cm} $ r_k^{main} \gets sample\left(\left\{r_k^1, r_k^2, ..., r_k^{L}\right\}\right)$ 
\STATE \hspace{0.5cm} $ \mathcal{L}_{unsup,k} \gets loss(r_k^{main}, \hat{y}_k^{final}) $ based on (\ref{equ:equ3})
\STATE \hspace{0.5cm} $ \mathcal{L} \gets \frac{1}{L} \sum_{j=1}^{L} \mathcal{L}_{sup,s}^j + \lambda \mathcal{L}_{unsup,k} $ 
\STATE \hspace{0.5cm} Minimize $\mathcal{L}$ to update model Q
\STATE {\textsc{\textbf{OUTPUT:}}} Trained Model Q
\end{algorithmic}
\end{algorithm}

Another form of perturbation involves the use of dropout layers in the proposed DiverseHead structure to increase the diversity of features during training. We introduce a dropout layer after each convolutional block in each segmentation head of the network, as shown in Figure \ref{fig:Two_perturbation_methods} (b). The pseudocode of this method is given in \textbf{Algorithm} \ref{alg:alg2}. As both dynamic freezing and dropout perturbation utilise the same semi-supervised learning framework, DiverseHead, the network initialisation and supervised loss calculation procedures stay the same. Using dropout in DiverseHead, specific components of the weights in the heads of the network are randomly assigned a value of zero with a dropout rate (determining the probability), using samples derived from a Bernoulli distribution. This dropout operation is employed to enhance the variability of the output predictions. In the conducted experiments, the dropout rate is set as a hyperparameter with a value of 0.5. Although the dropout-based approach adheres to the same training pipeline with dynamic freezing outlined above, the distinction is that all heads remain unfrozen. 

To evaluate the efficacy of individual perturbation methods in conjunction with the proposed DiverseHead techniques, each method is applied independently within the proposed framework. The performance of each combination is assessed in Section \ref{sec:results}. Considering that various datasets may require differing levels of diversity in pseudo labels within the proposed DiverseHead framework, adjustments can be made by altering the number of frozen heads and dropout rates in the methods of dynamic freezing and dropout, respectively.

\section{Cross-model Supervision (DiverseModel)}
\label{sec:proposed2}
\begin{figure}[!t]
\centering
\includegraphics[width=0.5 \textwidth]{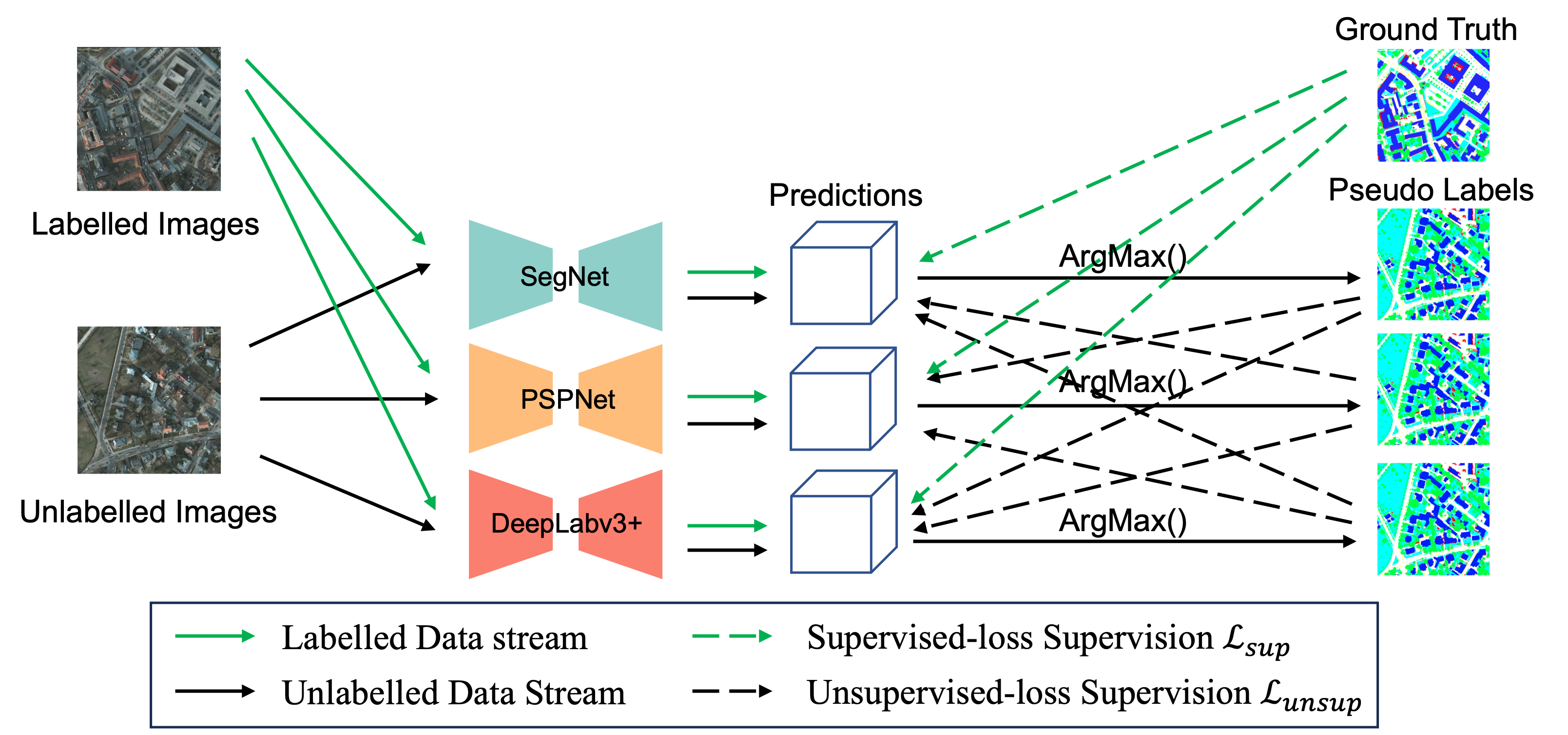}
\caption{DiverseModel: an online semi-supervised learning approach.}
\label{fig_2}
\end{figure}

\begin{figure*}[!t]
\centering
\includegraphics[width=0.8\textwidth]{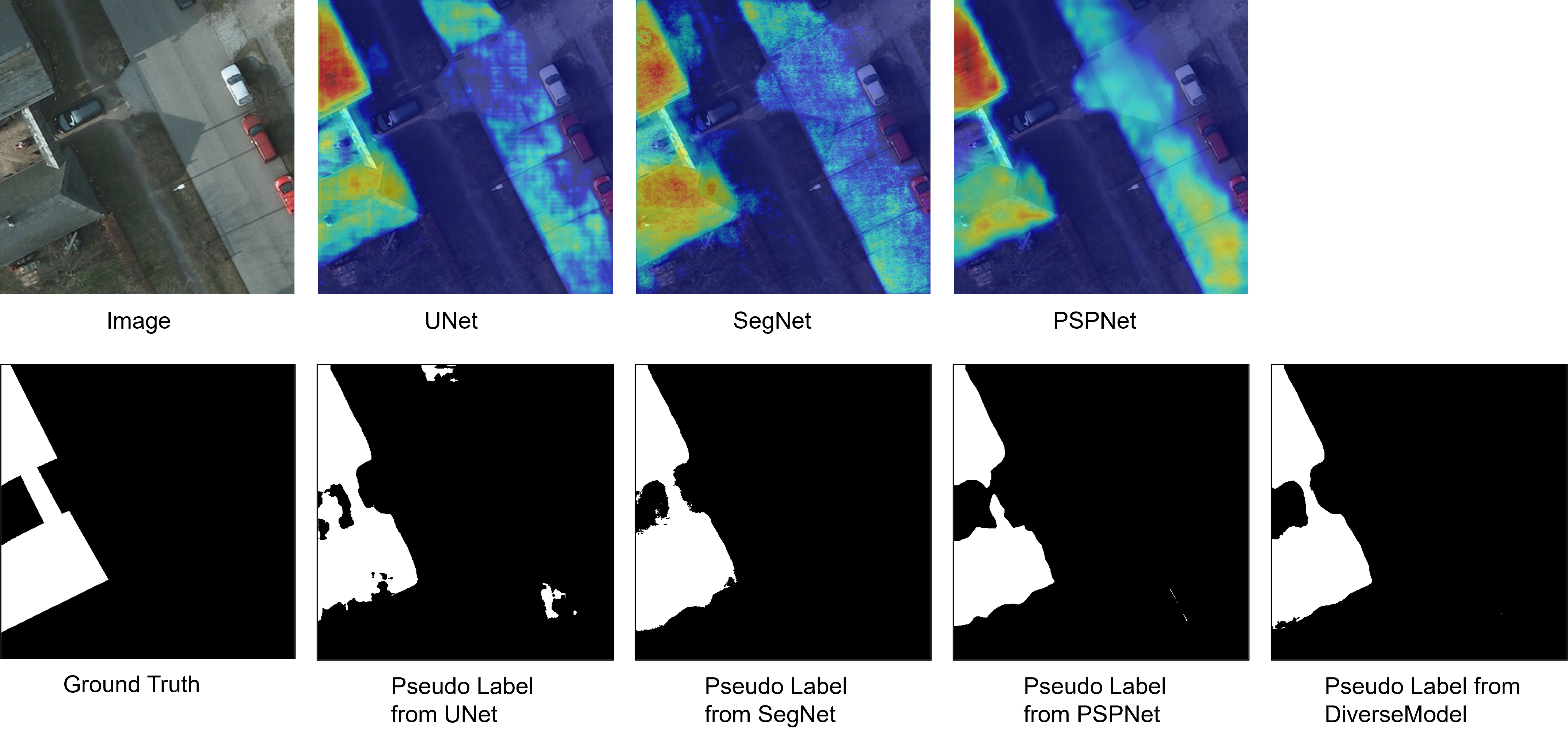}
\caption{The upper section displays Grad-CAM outputs from individual networks within the DiverseModel architecture using the Potsdam dataset. The lower section showcases both the ground truth and pseudo labels generated through predictions from each network. In the lower right corner, the pseudo-label from DiverseModel is presented.  }
\vspace{-10pt}
\label{fig_grad_cam}
\end{figure*}

The proposed DiverseModel differs from CPS, exploring different networks in parallel to generate comprehensive pseudo labels. As shown in Figure \ref{fig_2}, the DiverseModel structure includes three distinct networks, which can be various semantic segmentation networks. In this paper, we choose three widely used segmentation networks for experiments, which are PSPNet \cite{zhao2017pyramid}, UNet \cite{ronneberger2015u}, SegNet \cite{badrinarayanan2017segnet}. \textbf{Algorithm} \ref{alg:alg3} presents the pseudocode of the DiverseModel method. Since different networks pay different and complementary attention to the same input, this offers the basis for they are able to benefit from each other. In order to provide evidence for this claim, we executed the Gradient-weighted Class Activation Mapping (Grad-CAM) \cite{selvaraju2017grad} technique for every network employed within the framework of the DiverseModel architecture. Grad-CAM visualises the areas of an image that are important to the model predictions from each network. Figure \ref{fig_grad_cam} depicts an example grad-CAM analysis for the BUILDING class in the Potsdam data set. Examining Figure \ref{fig_grad_cam}, we can see that different networks pay different and complementary attention to the same input, and the pseudo labels from the DiverseModel prediction show the highest quality.

The labelled data is used in a regular supervised learning manner to train these models by using the standard cross-entropy loss function $\ell_{ce}$. The supervised loss $\mathcal{L}_{sup, s}$ is expressed as:

\begin{equation}
\label{equ:equ5}
\mathcal{L}_{sup, s} = \frac{1}{3}\sum_{n=1}^{3} \frac{1}{W \times H} \sum_{i=1}^{W \times H}\ell_{c e}\left(p_{i,s}^n, y_{i,s}\right),
\end{equation}
where $p_{i,s}^n$ represents the $i~th$ predicted pixel of the $s~th$ sample from the $n~th$ network.

In addition, unlabelled data is used to generate pseudo labels, which are then exploited for cross-supervision to inform each network. Different from the version presented in \cite{ma2023confidence}, all loss calculations in this work solely focus on cross-entropy loss to avoid the variations in performance resulting from different types of loss functions. The predictions obtained by each network are denoted as $\{p^1,p^2,p^3\}$, which are used for generating pseudo labels $\{\hat{r}^1, \hat{r}^2, \hat{r}^3\}$ through the $argmax$ operation. For instance, the cross pseudo supervision loss $\mathcal{L}_{unsup}^{12}$ between the prediction $p^1$ from the first network and the pseudo label $r^2$ generated by the second network is defined as:

\begin{equation}
\label{equ:equ6}
\mathcal{L}_{unsup,k}^{12} = \frac{1}{W \times H} \sum_{i=1}^{W \times H}\ell_{c e}\left(p_{i,k}^1, \hat{r}_{i,k}^2\right).
\end{equation}
where, $p_{i,k}^1$ represents the $i~th$ predicted pixel of the $k~th$ sample from the $1~st$ network.

\begin{algorithm}[h]
\small
\caption{\small DiverseModel Semi-supervised Learning Pseudocode. The Labelled Training Dataset is defined as $\mathcal{B}^l$. Since the number of labelled data is smaller than that of unlabelled data, the labelled data is used in cycles for one epoch. We define it as $\textit{cycle} \left(\mathcal{B}^l\right)$.}\label{alg:alg3}
\begin{algorithmic}
\STATE {\textsc{\textbf{Initialization:}}} 
\STATE \hspace{1.3cm} Randomly initialise three models, PSPNet $Q^{1}$,
\STATE \hspace{1.3cm} UNet $Q^{2}$, SegNet $Q^{3}$
\STATE {\textsc{\textbf{INPUT:}}} Labelled Training Dataset $\mathcal{B}^l=\left\{\left(x_i, y_i\right)\right\}_{i=1}^M$
\STATE \hspace{1.4cm} Unlabelled Training Dataset $\mathcal{B}^u=\left\{u_i\right\}_{i=1}^N$ 
\STATE \textbf{for} $ \left\{\left(x_s, y_s\right),u_k\right\}_{k=1, s=k\%M}^N \in \left\{\textit{cycle} \left(\mathcal{B}^l\right) , \mathcal{B}^u\right\}$ \textbf{do}
\STATE \hspace{0.5cm} $ \mathcal{L}_{sup,s} = loss(Q^{1}(x_s), y_s) $ + $ loss(Q^{2}(x_s), y_s) $ 
\STATE \hspace{1.7cm} + $loss(Q^{3}(x_s), y_s) $ based on (\ref{equ:equ1})
\STATE \hspace{0.5cm} $ \hat{r}_k^{1} \gets argmax(Q^{3}(u_k))$
\STATE \hspace{0.5cm} $ \hat{r}_k^{2} \gets argmax(Q^{3}(u_k))$
\STATE \hspace{0.5cm} $ \hat{r}_k^{3} \gets argmax(Q^{3}(u_k))$
\STATE \hspace{0.5cm} $ \mathcal{L}_{unsup,k} = loss(Q^{1}(u_k), \hat{r}_k^{2}) $ + $ loss(Q^{1}(u_k), \hat{r}_k^{3}) $ 
\STATE \hspace{1.7cm} + $loss(Q^{2}(u_k), \hat{r}_k^{1}) $ + $loss(Q^{2}(u_k), \hat{r}_k^{3}) $ 
\STATE \hspace{1.7cm} + $loss(Q^{3}(u_k), \hat{r}_k^{1}) $ + $loss(Q^{3}(u_k), \hat{r}_k^{2}) $ 
\STATE \hspace{0.5cm} $ \mathcal{L} \gets \mathcal{L}_{sup,s} + \lambda \mathcal{L}_{unsup,k} $ 
\STATE \hspace{0.5cm} Minimize $\mathcal{L}$ to update model $Q^{1}$, $Q^{2}$, $Q^{3}$
\STATE {\textsc{\textbf{OUTPUT:}}} Trained Model $Q^{1}$, $Q^{2}$, $Q^{3}$
\end{algorithmic}
\end{algorithm}

The cross-pseudo supervision among the three networks creates six losses in the same way. The unsupervised loss $\mathcal{L}_{unsup}$ is the average of the six individual losses, as shown below

\begin{equation}
\begin{aligned}
\mathcal{L}_{unsup,k} &= \frac{1}{6} \Big( \mathcal{L}_{unsup,k}^{12} + \mathcal{L}_{unsup,k}^{13} + \mathcal{L}_{unsup,k}^{21} \\
&\quad + \mathcal{L}_{unsup,k}^{23} + \mathcal{L}_{unsup,k}^{31} + \mathcal{L}_{unsup,k}^{32} \Big).
\end{aligned}
\end{equation}
The total loss $\mathcal{L}$ is the linear addition of $\mathcal{L}_{sup,s}$ and $\mathcal{L}_{unsup,k}$, which is previously given in equation (\ref{equ:equ4})

\section{Dataset Description}
\label{sec:Datasets}
We employed diverse remote sensing datasets to assess both the proposed techniques and state-of-the-art methods, specifically including (1)~the ISPRS Potsdam dataset \cite{rottensteiner2012isprs}, (2)~the DFC2020 dataset \cite{robinson2021global}, (3)~the RoadNet dataset \cite{liu2018roadnet}, and (4)~the Massachusetts Buildings dataset \cite{MnihThesis}. In the sequel, we share the details of each dataset utilised in this paper.

\textit{1) ISPRS Potsdam} Semantic Labelling dataset is an open-access benchmark dataset provided by the International Society for Photogrammetry and Remote Sensing (ISPRS). The true orthophoto (TOP) and DSM modalities have a ground sampling distance of 5 cm. Six land cover classes were identified by hand annotation of this dataset: \textit{impervious surfaces, buildings, low vegetation, trees, cars}, and \textit{clutter/background}. This dataset has 38 patches, each measuring $6000 \times 6000$ pixels. The patches include orthorectified optical pictures of red, green, and blue bands, as well as infrared (IR) and matching digital surface models (DSM). All of these data tiles were divided into $512\times512$ patches for computational reasons, yielding 3456, 201, and 1815 samples for the training, validation, and test sets, respectively. We randomly select a quarter of the training data as labelled data and use the remaining three quarters as unlabelled data for all SSL approaches. 

\textit{2) DFC2020} is the 2022 IEEE GRSS Data Fusion Contest dataset, which is based on the SEN12MS dataset \cite{schmitt2019sen12ms}. It provides Sentinel-1 SAR imagery, Sentinel-2 multispectral imagery, and corresponding land cover maps with ten coarser-grained classes. The size of all patches is \mbox{$256\times256$ pixels}. There are 6112, 986 and 5127 images for training, validation and test sets, respectively. In this paper, one-fifth of the labelled data is for training, while the remaining four-fifths of the data is employed as unlabelled data. 

\textit{3) RoadNet} is a benchmark dataset for roadnet detection with 0.21-m spatial resolution.
It includes RGB images and related road surface maps for the segmentation task. The number of samples for training, validation and testing is 410, 45 and 387, respectively. A quarter of the annotated training data is used for the supervised part, while the remaining three-quarters of the training data are employed as unlabelled data for SSL training purposes.

\textit{4) Massachusetts} Buildings dataset predominantly encompasses urban and suburban regions, encompassing structures of varying scales. 
It consists of 151 aerial RGB images with a resolution of $1~m^2$ per pixel and corresponding building masks with the size of $1500\times1500$ pixels. The dataset was split into 137 training, 10 testing, and 4 validation images with labels. One-quarter of the annotated training images are used for supervision, while the remaining three-quarters are treated as unlabelled for SSL.

\section{Experimental analysis}
\label{sec:results}
\subsection{Implementation Details}
\label{Implementation_Details}
Our approaches are implemented using PyTorch. Following \cite{chen2018encoder}, we employed a polynomial learning rate policy with a mini-batch SGD optimizer, where the current learning rate is calculated as the initial learning rate multiplied by $\left(1-\frac{\text { iter }}{\text { max-iter }}\right)^{\text {power }}$. The initial learning rate and power are set to 0.01 and 0.9, respectively. All experiments were conducted on the GW4 Isambard with an NVIDIA A100-sxm GPU and an AMD EPYC 7543P CPU ~\cite{gw4isambard}.

For a fair comparison, both the state-of-the-art methods and the proposed DiverseHead use DeepLabv3+ with a ResNet-50 backbone pre-trained on ImageNet as the semantic segmentation network. We comprehensively analysed all methods by quantifying performance via class-related measures, including overall accuracy (OA), user’s accuracy (UA), \mbox{producer’s accuracy (PA)}, mean intersection over union (mIoU), and $F_1$-score. Expressions of all five performance metrics are given as follows: $\mathrm{OA}=\frac{TP+TN}{TP+TN+FP+FN}$, $\mathrm{UA}=\frac{TP}{TP+FP}$, $\mathrm{PA}=\frac{TP}{TP+FN}$,  $\mathrm{mIoU}=\frac{|\mathrm{TP}|}{|\mathrm{TP}+\mathrm{FN}+\mathrm{FP}|}$, $F_1=\frac{2 \cdot PA \cdot UA}{PA+UA}$, where \textit{TP}, \textit{TN}, \textit{FP}, and \textit{FN} refer to the numbers of pixels that are true positives, true negatives, false positives, and false negatives for each class, respectively. 

\subsection{Quantitative Results and Analysis}

We evaluate the proposed approaches, along with classic and state-of-the-art SSL methods, using the five performance metrics outlined above, across four remote sensing imagery segmentation datasets: Potsdam, DFC2020, RoadNet, and Massachusetts Building. The average results across four datasets are presented in Table \ref{tab:addlabe4}. It presents an overall performance of the proposed methods using a single network in comparison to traditional SSL methods that employ dual networks. Since some state-of-the-art SSL methods in computer vision, such as UniMatch, iMAS, and AugSeg, are designed for RGB image segmentation and rely on RGB-based datasets for pretraining, these methods are not included in the average performance across the four datasets, including multi-band remote sensing data. However, a detailed performance comparison of these state-of-the-art and classic SSL methods across various datasets, including both multi-band and RGB remote sensing imagery, will be presented and discussed later in this section. From the average results in Table \ref{tab:addlabe4}, the proposed DiverseHead (DF) delivers the best overall performance, achieving the highest score in 4 out of 5 metrics, which are highlighted in red. DiverseModel demonstrates the second-best performance across most of the metrics. In particular, for metrics of UA, the proposed DiverseModel exhibits an improvement of over 3.48\% compared to another network-perturbation-based approach, CPS. Although the performance of DiverseHead (DT) is slightly lower than that of DiverseHead (DF) and DiverseModel, these three methods show similar average performance, but both of them outperform the other compared methods. With these used remote sensing datasets, MT demonstrates notably inferior average performance across various metrics. CCT, GCT, and CPS exhibit comparable performance, yet the performance superiority of CPS is notably evident in its PA metric. It is important to highlight that DiverseHead is a very lightweight method and reaches a very competitive average performance compared to others.

\begin{table}[htbp]
  \centering
  \small
  \caption{Average performance comparison with the classic SSL methods with multiple networks on four datasets. DT and DF indicate dropout and dynamic freezing, respectively.} 
    \begin{tabular}{@{}l@{\kern.5em}l@{\kern.5em}c@{\kern.5em}c@{\kern.5em}c@{\kern.5em}c@{}}
    \toprule
       Models   & OA    & UA  & PA & mIoU  & $F_1$ \\
    \midrule
    MT \cite{tarvainen2017mean}  & 86.35\% &	74.93\% &	81.53\% &	66.03\% &	77.99\% \\
    CCT \cite{ouali2020semi} & 87.14\% &	76.49\% &	82.50\% &	67.76\% &	79.23\% \\
    GCT \cite{ke2020guided} &87.45\% &	75.35\% &	82.31\% &	67.19\% &	78.65\% \\
    CPS \cite{chen2021semi} &88.06\% &	75.97\% &	85.27\% &	68.23\% &	80.22\% \\
    
    DiverseModel \cite{ma2023confidence} & 88.77\% &	 \textcolor[rgb]{ 1,  0,  0}{\textbf{79.54\%}} &	85.18\% &	70.92\% &	82.02\% \\
    DiverseHead (DT) &88.69\% &	78.51\% &	85.42\% &	70.63\% &	81.64\% \\
    DiverseHead (DF) & \textcolor[rgb]{ 1,  0,  0}{\textbf{89.00\%}} &	79.14\% &	\textcolor[rgb]{ 1,  0,  0}{\textbf{85.83\%}} &	\textcolor[rgb]{ 1,  0,  0}{\textbf{71.28\%}} &	\textcolor[rgb]{ 1,  0,  0}{\textbf{82.17\%}}  \\
    \bottomrule
    \end{tabular}%
  \label{tab:addlabe4}%
\end{table}%


Table \ref{tab:addlabe5} presents the number of parameters of SSL frameworks during training. Apart from DiverseHead, all other classic SSL approaches typically involve parameters exceeding three hundred megabytes. Among these approaches, DiverseModel is the largest architecture; however, it outperforms other network-perturbation-based methods, including MT, CCT, GCT, and CPS, as demonstrated in Table \ref{tab:addlabe4}. In contrast, the proposed DiverseHead is highly lightweight, using just a single segmentation model with multiple heads, each consisting of only 2 convolutional layers, thus eliminating the need for multiple networks during training. Specifically, the parameter size of the DiverseHead (DT\&DF) is only 16\% bigger than that of the single network (Base in Table \ref{tab:addlabe5}), whereas the parameter size of other reference semi-supervised architectures is at least twice that of the single network. Despite requiring fewer parameters, the proposed DiverseHead method surpasses the classic SSL methods by at least 1\% in accuracy and 3\% in mIoU, while achieving performance comparable to the largest method, DiverseModel.

\begin{table}[htbp]
  \centering
  \small
  \caption{The required parameter size of each semi-supervised learning approach. Base means segmentation network DeepLabv3+. DT and DF present dropout and dynamic freezing, respectively.}
    \begin{tabular}{@{}c@{\kern0.5em}c@{\kern0.5em}c@{\kern0.5em}c@{\kern0.5em}c@{}}
    \toprule
            & CPS  \cite{chen2021semi}    & MT \cite{tarvainen2017mean}  & CCT  \cite{ouali2020semi}  &  GCT \cite{ke2020guided} \\ 
    \midrule
    Size (MB)   & 303.378   & 303.378  & 337.655  & 335.049\\
    \bottomrule
            &  \makecell[c]{DiverseModel \\ \cite{ma2023confidence}}     & \makecell[c]{DiverseHead \\ (DT)}  &  \makecell[c]{ DiverseHead \\ (DF)} &  Base \\
    \midrule
     Size (MB)  &  936.262  & 175.806  &  175.806 &   151.689 \\
    \bottomrule
    \end{tabular}%
  \label{tab:addlabe5}%
\end{table}%

Specifically, we present the results of different semi-supervised learning methods applied to the two multi-band remote sensing imagery datasets, namely Potsdam and DFC2020, in the table \ref{tab:detail_results1}. The performance of the \textit{DiverseNet}, namely DiverseHead and DiverseModel, is noticeably superior to that of the other listed semi-supervised learning methods across these evaluated datasets. In particular, for the Potsdam datasets, the proposed DiverseHead (DF) attains the highest performance across 4 out of 5 segmentation metrics, whilst DiverseModel emerges as the second-best method based on its results. However, in the case of the DFC2020 dataset, DiverseModel reaches the best for most performance metrics, securing the top position whilst DiverseHead closely follows as the second-best method, despite their similar performance. Images from Potsdam consist of 4 bands, while images from DFC2020 contain 13 bands. The DiverseModel exhibits greater proficiency in processing datasets with more bands, while Diverse is more effective in handling images with fewer bands. 

\begin{table*}
 \setlength\tabcolsep{3pt} 
  \centering
  \small
  \caption{Performance comparison with the state-of-the-art on datasets containing images with more than three bands, namely Potsdam and DFC2020. DT and DF present Dropout and dynamic freezing, respectively.}
    \begin{tabular}{c|p{3.5cm}p{1.75cm}p{1.75cm}p{1.75cm}p{1.75cm}p{1.5cm}}
    \toprule
          & Model & OA    & UA & PA & mIoU  & $F_1$ \\
    \toprule
    \multirow{7}[2]{*}{\begin{turn}{90}Potsdam\end{turn}} 
    &MT\cite{tarvainen2017mean} & 81.98\% & 73.66\% & 78.39\% & 63.07\% & 75.95\% \\
    &CCT\cite{ouali2020semi} & 82.66\% &	74.64\% &	77.62\% &	64.16\% &	76.10\% \\
    &GCT\cite{ke2020guided}  & 83.99\% & 75.65\% & 80.81\% & 65.80\% & 78.14\% \\
    &CPS\cite{chen2021semi} & 85.00\% & 75.76\% & \underline{82.94\%} & 66.69\% & 79.19\% \\
    &DiverseModel\cite{ma2023confidence} & \underline{85.76\%} & 76.75\% & \textcolor[rgb]{ 1,  0,  0}{\textbf{83.45\%}} & \underline{67.85\%} & \underline{79.96\% }\\
   
    &DiverseHead (DT)  & 84.66\% &	\underline{77.04\%} &	80.78\% &	67.12\% &	78.87\% \\
    &DiverseHead (DF) & \textcolor[rgb]{ 1,  0,  0}{\textbf{85.98\%}} &	\textcolor[rgb]{ 1,  0,  0}{\textbf{79.15\%}} &	82.87\% &	\textcolor[rgb]{ 1,  0,  0}{\textbf{69.63\%}} &	\textcolor[rgb]{ 1,  0,  0}{\textbf{80.97\%}}  \\
    \midrule
    \multirow{6}[2]{*}{\begin{turn}{90}DFC2020\end{turn}}
    &MT\cite{tarvainen2017mean} & 78.64\% & 59.59\% & 73.57\% & 50.44\% & 65.85\% \\
    &CCT\cite{ouali2020semi} & 79.71\% &	59.87\% &	76.40\% &	51.04\% &	67.13\% \\
    &GCT\cite{ke2020guided}   & 80.84\% & 61.47\% & 71.43\% & 52.17\% & 66.07\% \\
    &CPS\cite{chen2021semi} & 81.49\% &	61.74\% &	79.46\% &	53.20\% &	69.49\% \\
    &DiverseModel\cite{ma2023confidence} &   \underline{81.87\%} &	\textcolor[rgb]{ 1,  0,  0}{\textbf{62.20\%}} &	\textcolor[rgb]{ 1,  0,  0}{\textbf{80.69\%}} &	\underline{53.71\%} &	\textcolor[rgb]{ 1,  0,  0}{\textbf{70.25\%}}  \\
    &DiverseHead (DT) &  \textcolor[rgb]{ 1,  0,  0}{\textbf{82.02\%}} &	\underline{62.13\%} &	 \underline{80.61\%} &	\textcolor[rgb]{ 1,  0,  0}{\textbf{53.81\%}} &	\underline{70.18\%} \\
    
    &DiverseHead (DF) &  81.78\% &	61.81\% &	80.21\% &	53.46\% &	69.82\% \\
    
    \bottomrule
    \end{tabular}%
  \label{tab:detail_results1}%
\end{table*}%

\begin{table*}
 \setlength\tabcolsep{3pt} 
  \centering
  \small
  \caption{Performance comparison with the state-of-the-art on RGB-band datasets, namely Roadnet and Massachusetts. DT and DF present Dropout and dynamic freezing, respectively.}
    \begin{tabular}{c|p{5cm}p{1.75cm}p{1.75cm}p{1.75cm}p{1.75cm}p{1.5cm}}
    \toprule
          & Model & OA    & UA & PA & mIoU  & $F_1$ \\
    \toprule
    \multirow{12}[2]{*}{\begin{turn}{90}RoadNet\end{turn}}
    &MT\cite{tarvainen2017mean} &94.61\% &	86.87\% &	88.18\% &	79.04\% &	87.52\% \\
    &CCT\cite{ouali2020semi}  & 95.58\% &	88.67\% &	90.74\% &	82.18\% &	89.69\%  \\
    &GCT\cite{ke2020guided}   & 95.29\% &	85.53\% &	91.99\% &	80.35\% &	88.64\%  \\
    &CPS\cite{chen2021semi} & 95.81\% &	89.10\% &	91.36\% &	82.96\% &	90.21\%  \\
    &FixMatch \cite{sohn2020fixmatch}	& 96.02\% &	89.91\% &	91.63\% &	83.81\% &	90.76\% \\
    &UniMatch \cite{yang2023revisiting}	& 95.93\% &	90.58\% &	90.79\% &	83.71\% &	90.69\% \\
    &ICNet \cite{sohn2020fixmatch}	&  97.15\% &	 93.48\% &	 93.52\% &	 88.19\% &	 93.50\% \\
    &iMAS \cite{zhao2023instance} & 97.15\% & 94.15\% & 93.03\% & 88.32\% & 93.59\%\\
    &AugSeg \cite{zhao2023augmentation} & 97.36\% & 94.26\% & 93.79\% & 89.06\% & 94.03\%\\
    &DWL \cite{huang2024decouple} &  \underline{97.88\%} &  94.81\% & \textcolor[rgb]{ 1,  0,  0}{\textbf{95.48\%}} &  \underline{90.96\%} &  \underline{95.14\%}\\
    \cdashline{2-7}
    &DiverseModel\cite{ma2023confidence} & 96.84\% &	93.12\% &	92.58\% &	87.11\% &	92.84\% \\
    &  DiverseHead(DT) w/ Sngl-model & 96.85\% &	92.14\% &	93.31\% &	86.91\% &	92.72\%   \\
    &  DiverseHead(DF) w/ Sngl-model& 96.81\% &	92.60\% &	92.80\% &	86.88\% &	92.70\%  \\
    & DiverseHead(DT) w/ iMAS &97.47\%&94.24\%&94.20\%&89.40\%&94.22\%\\
    & DiverseHead(DT) w/ AugSeg  &97.70\%&\textcolor[rgb]{ 1,  0,  0}{\textbf{95.85\%}}&93.96\%& 90.50\%& 94.90\%\\
    & DiverseHead(DT) w/ DWL &  \textcolor[rgb]{ 1,  0,  0}{\textbf{97.96\%}} &  \underline{95.33\%} & \underline{95.36\%} &  \textcolor[rgb]{ 1,  0,  0}{\textbf{91.32\%}} &  \textcolor[rgb]{ 1,  0,  0}{\textbf{95.35\%}}\\
    \midrule
    \multirow{12}[2]{*}{\begin{turn}{90}Massachusetts\end{turn}} 
    &MT\cite{tarvainen2017mean}  & 90.16\% &	79.58\% &	85.97\% &	71.55\% &	82.65\% \\
    &CCT\cite{ouali2020semi}  & 90.59\% &	82.77\% &	85.23\% &	73.67\% &	83.98\% \\
    &GCT\cite{ke2020guided} & 89.67\% &	78.75\% &	85.01\% &	70.42\% &	81.76\% \\
    &CPS\cite{chen2021semi} & 89.95\% &	77.26\% &	87.33\% &	70.07\% &	81.98\%  \\
    &FixMatch \cite{sohn2020fixmatch}& 90.00\%	& 81.34\%	&	84.41\%	&	72.13\%	&	82.85\%	\\
    &UniMatch \cite{yang2023revisiting}& 89.86\%	& 79.95\%	&	84.79\%	&	71.28\%	&	82.30\%	\\
    &ICNet \cite{sohn2020fixmatch}	&  92.24\% &	  84.81\% &	  88.64\% &	 77.34\% &	 86.68\% \\
    &iMAS \cite{zhao2023instance}&92.40\%&83.89\%&89.90\%&77.26\%&86.79\%\\
    &AugSeg \cite{zhao2023augmentation} &91.27\%&80.92\%&88.68\%&73.98\%&84.62\%\\
    &DWL \cite{huang2024decouple} &  91.92\% &  \underline{88.29\%} & 86.00\% &  77.98\% &  87.13\%\\
    \cdashline{2-7}
    &DiverseModel\cite{ma2023confidence} & 90.62\% &	86.07\% &	84.01\% &	75.00\% &	85.03\% \\
    & DiverseHead(DT) w/ Sngl-model & 91.21\% &	82.73\% &	86.98\% &	74.67\% &	84.80\% \\
    & DiverseHead(DF) w/ Sngl-model & 91.43\% &	83.00\% &	87.45\% &	75.16\% &	85.17\% \\
    & DiverseHead(DT) w/ iMAS &\underline{92.92\%}&85.43\%&\textcolor[rgb]{ 1,  0,  0}{\textbf{90.27\%}}&78.84\%&87.78\%\\
    &DiverseHead(DT) w/ AugSeg  &\textcolor[rgb]{ 1,  0,  0}{\textbf{93.35}}\%&87.60\%&\underline{89.93\%}&\textcolor[rgb]{ 1,  0,  0}{\textbf{80.47\%}}&\textcolor[rgb]{ 1,  0,  0}{\textbf{88.75\%}}\\
    & DiverseHead(DT) w/ DWL &  92.57\% &  \textcolor[rgb]{ 1,  0,  0}{\textbf{88.49\%}} & 87.37\% &  \underline{79.25\%} &  \underline{87.93\%}\\
    \bottomrule
    \end{tabular}%
  \label{tab:detail_results2}%
\end{table*}%

The proposed DiverseHead with two perturbation methods is a foundational framework, requiring fewer parameters and making it highly suitable for users with limited computational resources. Also, this proposed foundational framework can be easily combined with various teacher-student SSL methods to improve their performance. Table \ref{tab:detail_results2} presents the results of the proposed methods and various SSL methods, including state-of-the-art methods in computer vision, which incorporate various enhancement strategies, on RGB remote sensing image segmentation datasets, specifically RoadNet and Massachusetts Building. DiverseHead demonstrates superior performance compared to other classic frameworks, such as MT, CCT, GCT, CPS,  FixMatch and UniMatch, while maintaining a lower parameter requirement and quicker training process. Although some currently proposed SSL methods, iMAS, AugSeg and DWL, outperform the proposed DiverseHead, integrating these state-of-the-art methods with the proposed DiverseHead idea leads to further performance improvement and reaches the best result in Table \ref{tab:detail_results2}.

\begin{figure}[t]
    \centering
    \begin{tikzpicture}
    \begin{axis}[
        ybar,
        bar width=9pt,
        width=\linewidth,
        height=5cm,
        enlarge x limits=0.08,
        ylabel={Training Time (s)},
        xlabel={},
        symbolic x coords={MT, CCT, GCT, CPS, FixMatch, UniMatch, iMAS, AugSeg, DiverseHead},
        xtick=data,
        x tick label style={rotate=45, anchor=east},
        ymin=0,
        ymax=230,
        ytick={0,50,100,150,200},
        axis line style={-},
        axis x line*=bottom,
        axis y line*=left,
        tick style={draw=none},
        nodes near coords,
        every node near coord/.append style={font=\small, rotate=90, anchor=west, text=black},
    ]
    \addplot+[ybar, fill={rgb,255:red,255;green,195;blue,207},draw=none] coordinates {
        (MT,63.96) 
        (CCT,120.78) 
        (GCT,206.40) 
        (CPS,93.80) 
        (FixMatch,87.83) 
        (UniMatch,104.08) 
        (iMAS,152.18) 
        (AugSeg,81.72)
        (DiverseHead,40.65)
    };
    \end{axis}
    \end{tikzpicture}
    \vspace{-10pt}
    \caption{Training time (in seconds) for 100 iterations across different SSL methods. DiverseHead refers to the dropout version of our proposed method.}\vspace{-20pt}
    \label{fig:train_time}
\end{figure}

To further support the computational efficiency of the proposed method, we measured the wall-clock training time required for 100 iterations across various state-of-the-art SSL frameworks, using identical hardware and implementation settings. As shown in Figure~\ref{fig:train_time}, \textit{DiverseHead (dropout)} achieves the lowest training time (40.65 seconds) among all compared methods. This demonstrates that our approach introduces minimal computational overhead, providing strong empirical evidence of its lightweight nature during training.

In order to further prove the efficiency of multi-head supervision, we considered a downgraded version of the current DiverseHead approach, called single-head supervision (SHS), which consists of only a single head. During each training iteration, the current model's predictions for unlabelled data are used to generate pseudo labels through an $argmax$ operation. These generated pseudo labels are then used to calculate the unsupervised loss. Also, we provide the performance of the segmentation network namely DeepLabv3+, called Base in Table \ref{tab:addlabe3}, only trained by labelled data (part of each dataset) without using unlabelled data. The performance of the Base and SHS, along with the proposed DiverseHead, are presented in Table \ref{tab:addlabe3} to provide experimental evidence of the improvement and efficiency of the use of multiple heads in the DiverseHead architecture. While the single-head supervision benefits from the pseudo labels and exhibits better performance compared to the Base model, especially improving the metric of PA, both versions of DiverseHead approaches demonstrate a further significant improvement by taking advantage of multiple heads. Specifically, the mIoU of DiverseHead (DF) is 6.91\% higher than that of Base and 5.12\% higher than that of SHS for Potsdam. While the performance of DiverseHead (DT) is slightly lower than that of DiverseHead (DF), its performance is still much better than that of SHS and Base. Similarly, Both DiverseHead versions demonstrate superior performance on the RoadNet dataset compared to SHS and Base, particularly evident in the mIoU metric, where DiverseHead (DT) exhibits a significant improvement of 6.8\% over the Base. 

\begin{figure}
\centering

\includegraphics[width=0.5\textwidth]{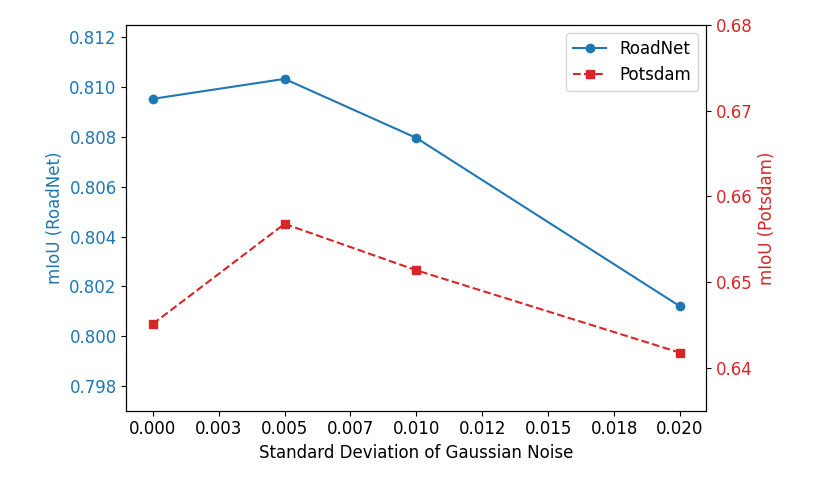}
\vspace{-15pt}
\caption{Variation in mIoU as the standard deviation of the Gaussian noise added for input perturbation increases}
\vspace{-12pt}
\label{fig:Noise_ablation_input_perturbation}
\end{figure}

To evaluate the effectiveness of the proposed perturbation methods, DT and DF, in comparison to input perturbation, we conducted supplementary experiments involving data perturbation by introducing Gaussian noise to the input images. Consistency regularisation is achieved by enforcing the network to produce consistent predictions for the original and noisy input images, which is termed Input Perturbation in Table \ref{tab:addlabe3}. Both versions of DiverseHead exhibit better performance than Input Perturbation, whose performance is similar to SHS. 
To use the optimal Hyperparameter applied for input perturbation for fair comparison, hyperparameter tuning experiments are conducted on two datasets to confirm the standard deviation (SD) of Gaussian noise. Figure \ref{fig:Noise_ablation_input_perturbation} illustrates the variation in mIoU as the SD changes. The mIoU value for both Potsdam and Roadnet peaked at an SD of 0.01. Thus, we set the SD of Gaussian noise for input perturbation 0.01 for the Potsdam and RoadNet datasets.

\begin{table}
 \setlength\tabcolsep{3pt} 
  \centering
  \small
  \caption{Performance comparison with single-head supervision (SHS) and baseline model which is only supervised by labelled data. DT and DF present dropout and dynamic freezing, respectively.}
    \begin{tabular}{@{}c|l@{\kern.5em}c@{\kern.5em}c@{\kern.5em}c@{\kern.5em}c@{\kern.5em}c@{}}
    \toprule
          & Model & OA    & UA & PA & mIoU  & $F_1$ \\
    \midrule
    \multirow{3}[2]{*}{\begin{turn}{90}Potsdam\end{turn}}
    &Base  & 81.64\% & 73.86\% & 76.05\% & 62.72\% & 74.94\% \\
    &SHS & 83.43\% & 74.57\% & 79.93\% & 64.51\% & 77.16\% \\
    &Input Perturbation & 83.89\% & 75.73\% & 80.54\% & 65.69\% & 78.06\% \\
    &DiverseHead (DT)  & 84.66\% &	77.04\% &	80.78\% &	67.12\% &	78.87\% \\
    &DiverseHead (DF) & \textcolor[rgb]{ 1,  0,  0}{\textbf{85.98\%}} &	\textcolor[rgb]{ 1,  0,  0}{\textbf{79.15\%}} &	\textcolor[rgb]{ 1,  0,  0}{\textbf{82.87\%}} &	\textcolor[rgb]{ 1,  0,  0}{\textbf{69.63\%}} &	\textcolor[rgb]{ 1,  0,  0}{\textbf{80.97\%}}  \\
    
    \midrule
    \multirow{3}[2]{*}{\begin{turn}{90}RoadNet\end{turn}}
    &Base  & 95.17\% &	87.62\% &	89.81\% &	80.73\%	& 88.71\%  \\
    &SHS &  95.44\%	& 87.31\% &	91.17\%	& 81.37\%	& 89.20\%    \\
    & Input Perturbation & 95.31\% &87.42\% &	90.52\% &	81.03\% &	88.94\%  \\
        &DiverseHead (DT) & \textcolor[rgb]{ 1,  0,  0}{\textbf{96.85\%}} &	92.14\% &	\textcolor[rgb]{ 1,  0,  0}{\textbf{93.31\%}} &	\textcolor[rgb]{ 1,  0,  0}{\textbf{86.91\%}} &	\textcolor[rgb]{ 1,  0,  0}{\textbf{92.72\%}}   \\
    &DiverseHead (DF)  & 96.81\% &	\textcolor[rgb]{ 1,  0,  0}{\textbf{92.60\%}} &	92.80\% &	86.88\% &	92.70\%  \\
    \bottomrule
    \end{tabular}%
  \label{tab:addlabe3}%
\end{table}%

\begin{table}
 \setlength\tabcolsep{3pt} 
  \centering
  \small
  \caption{Ablation study on the effect of head number in DiverseHead with dynamic freezing.}
    \begin{tabular}{@{}c@{}|c@{\kern.5em}c@{\kern.5em}c@{\kern.5em}c@{\kern.5em}c@{\kern.5em}c@{}}
    \toprule
          & \# of heads & OA    & UA & PA & mIoU  & $F_1$ \\
    \midrule

    \multirow{2}[2]{*}{\begin{turn}{90}Potsdam\end{turn}}
    & 8 & 85.92\% &	77.86\% &	83.86\% &	68.91\% &	80.75\%  \\
    & 10 & 85.98\% &	79.15\% &	82.87\% &	69.63\% &	80.97\%  \\
    & 12 & 85.70\% &	78.54\% &	82.55\% &	68.96\% &	80.50\%  \\
    \midrule
    \multirow{2}[2]{*}{\begin{turn}{90}RoadNet\end{turn}}
    & 8  &96.63\% &	92.15\% &	92.42\% &	86.21\% &	92.29\%  \\
    & 10 &96.81\% &	92.60\% &	92.80\% &	86.88\% &	92.70\% \\
    & 12 &96.80\% &	92.59\% &	92.77\% &	86.84\% &	92.68\%   \\
    \midrule
    \multirow{2}[2]{*}{\begin{turn}{90}Average\end{turn}} 
    & 8  &91.27\% &	85.01\% &	\textcolor[rgb]{ 1,  0,  0}{\textbf{88.14\%}} &	77.56\% &	86.52\%  \\
    & 10 &\textcolor[rgb]{ 1,  0,  0}{\textbf{91.39\%}} &	\textcolor[rgb]{ 1,  0,  0}{\textbf{85.88\%}} &	87.83\% &	\textcolor[rgb]{ 1,  0,  0}{\textbf{78.25}}\% &	\textcolor[rgb]{ 1,  0,  0}{\textbf{86.83\%}} \\ 
    & 12 &91.25\% &	85.57\% &	87.66\% &	77.90\% &	86.59\%\\ 
    \bottomrule
    \end{tabular}%
  \label{tab:addlabe6}%
\end{table}%

To investigate the influence of head count in the DiverseHead framework on performance, an ablation study was conducted on the Potsdam and RoadNet datasets using the proposed DiverseHead framework with dynamic freezing perturbation. The results in Table \ref{tab:addlabe6} show that using 10 heads for DiverseHead achieves the best average performance across most metrics, despite the small performance differences among the variants. This supports the choice of using 10 heads as an effective design decision.

To show the efficiency of DiverseModel, consisting of three different segmentation networks, we also evaluated the performance of each member network within the DiverseModel on the Potsdam dataset in Table \ref{tab:addlabe7}. The component models UNet, SegNet, and PSPNet use unlabelled data through the way of SHS. DiverseModel demonstrates a significant improvement across all performance metrics in the segmentation task compared to each network. In particular, the metric of PA experiences a notable improvement of 4.57\% compared to the best-performing individual component model. This phenomenon can be attributed to the enhancement of pseudo-label diversity through cross-model supervision, resulting in a significant improvement in PA (recall) during the test phase. The findings suggest that the cross-supervision of different networks has the potential to achieve superior performance compared to the best-performing individual component.

\begin{table}
  \centering
  \small
  \caption{Performance comparison of the DiverseModel with its constituent networks on the Potsdam dataset }
    \begin{tabular}{@{}c@{\kern.5em}c@{\kern.5em}c@{\kern.5em}c@{\kern.5em}c@{\kern.5em}c@{}}
    \toprule
         Model & OA    & UA  & PA & mIoU  & $F_1$ \\
    \midrule
    UNet  & 83.28\% & 74.18\% & 78.88\% & 64.51\% & 76.46\% \\
    SegNet & 82.37\% & 73.53\% & 77.46\% & 63.24\% & 75.45\% \\
    PSPNet & 81.96\% & 74.23\% & 76.99\% & 63.08\% & 75.58\% \\
    DiverseModel & \textcolor[rgb]{ 1,  0,  0}{\textbf{85.76\%}} & \textcolor[rgb]{ 1,  0,  0}{\textbf{76.75\%}} & \textcolor[rgb]{ 1,  0,  0}{\textbf{83.45\%}} & \textcolor[rgb]{ 1,  0,  0}{\textbf{67.85\%}} & \textcolor[rgb]{ 1,  0,  0}{\textbf{79.96\%}} \\
    \bottomrule
    \end{tabular}%
    \vspace{-10pt}
  \label{tab:addlabe7}%

\end{table}%

\subsection{Qualitative Results and Analysis}
\begin{figure*}
\centering \setlength \tabcolsep{3pt}
    \begin{tabular}{cm{0.1\linewidth}m{0.1\linewidth}m{0.1\linewidth}m{0.1\linewidth}m{0.02\linewidth}m{0.1\linewidth}m{0.1\linewidth}m{0.1\linewidth}m{0.1\linewidth}c}
        & \centering Potsdam & \centering DFC2020 & \centering RoadNet & \centering \makecell[l]{Massa- \\ chusetts }  & &\centering Potsdam & \centering DFC2020 & \centering RoadNet & \centering \makecell[l]{Massa- \\ chusetts }  & \\ \vspace{-3pt}
         (1) & \includegraphics[width=\linewidth]{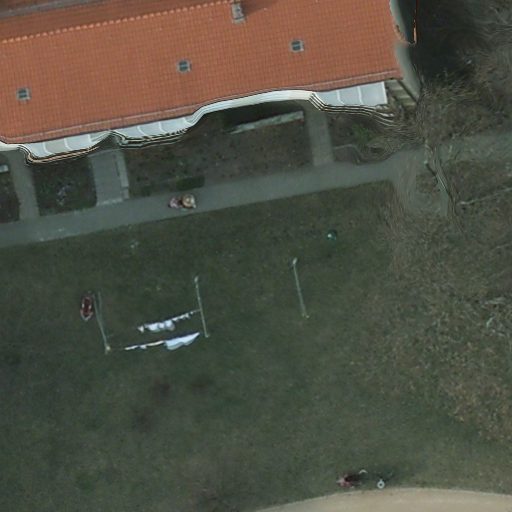} & \includegraphics[width=\linewidth]{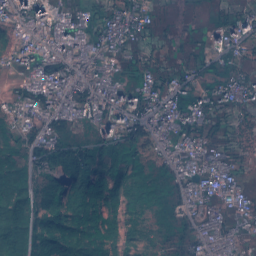} & \includegraphics[width=\linewidth]{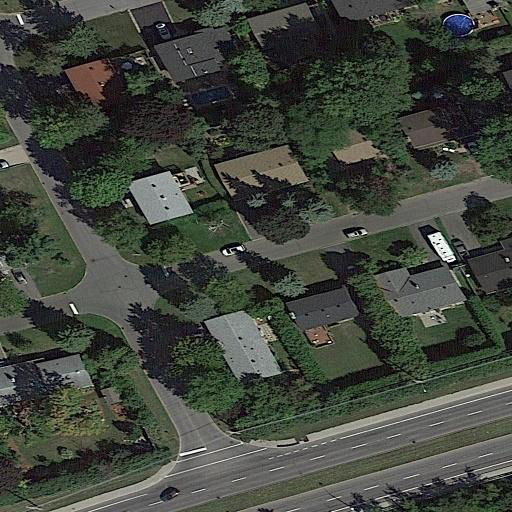} & \includegraphics[width=\linewidth]{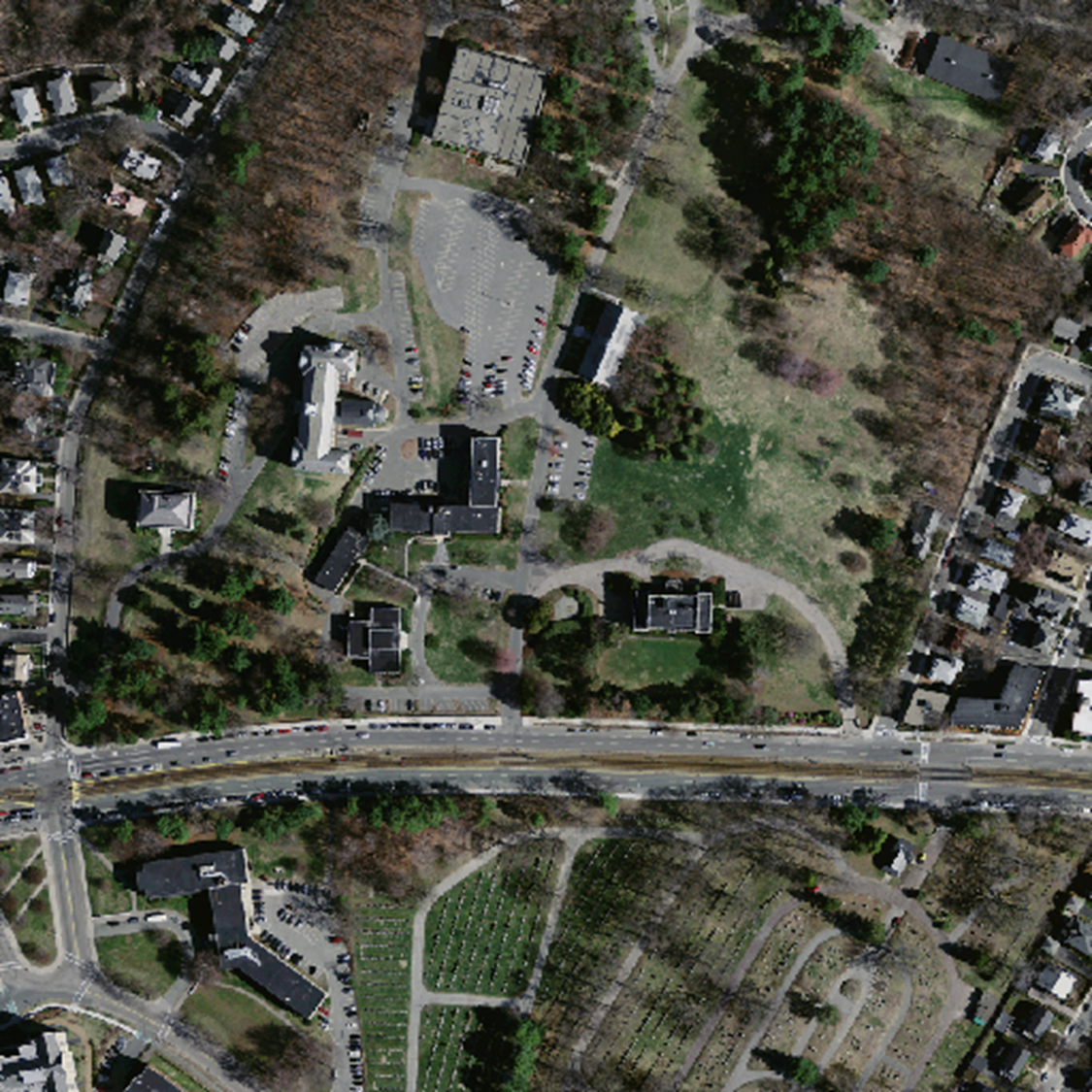} & &
        \includegraphics[width=\linewidth]{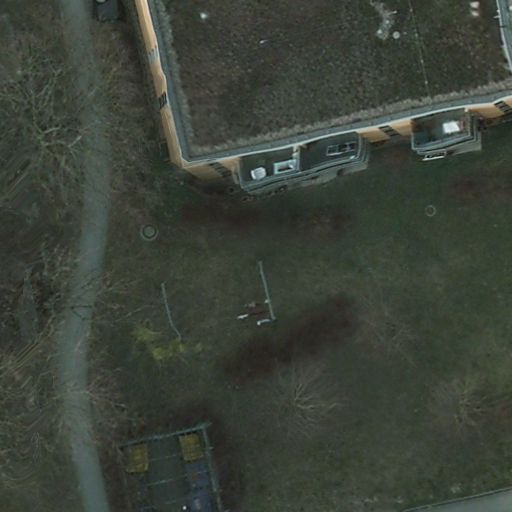} & \includegraphics[width=\linewidth]{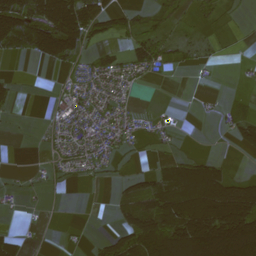} & \includegraphics[width=\linewidth]{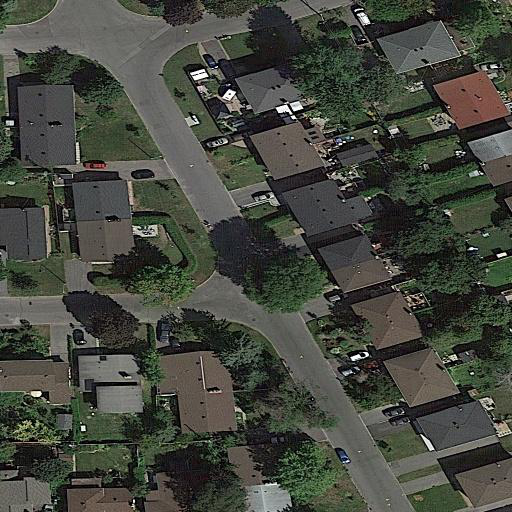} & \includegraphics[width=\linewidth]{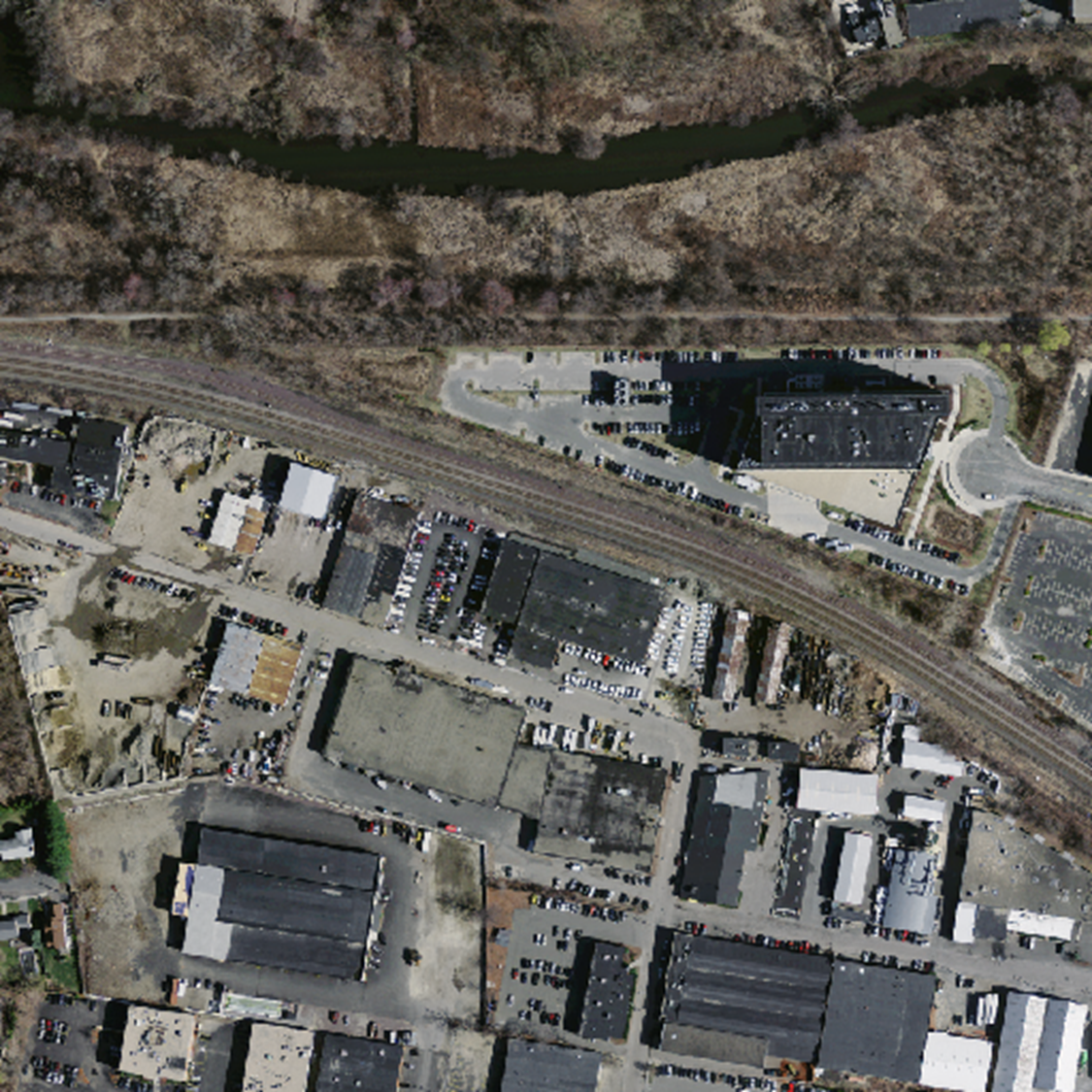} & \\

        \vspace{-3pt}(2) & \includegraphics[width=\linewidth]{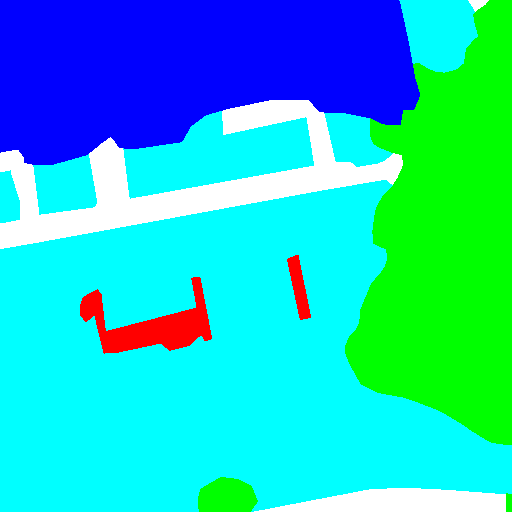} & \includegraphics[width=\linewidth]{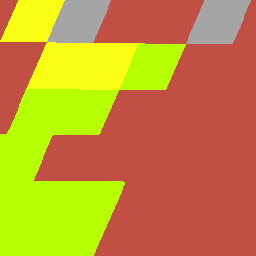} & \includegraphics[width=\linewidth]{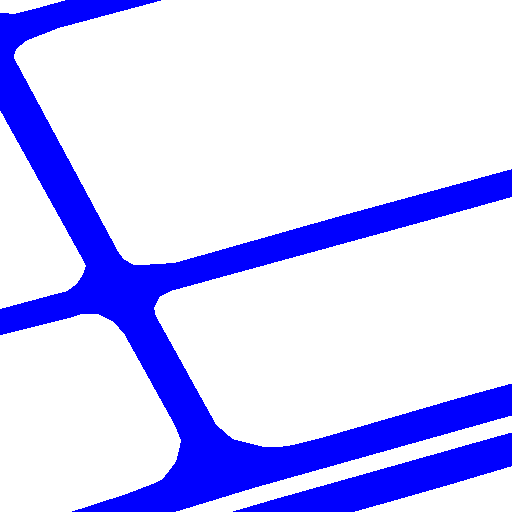} & \includegraphics[width=\linewidth]{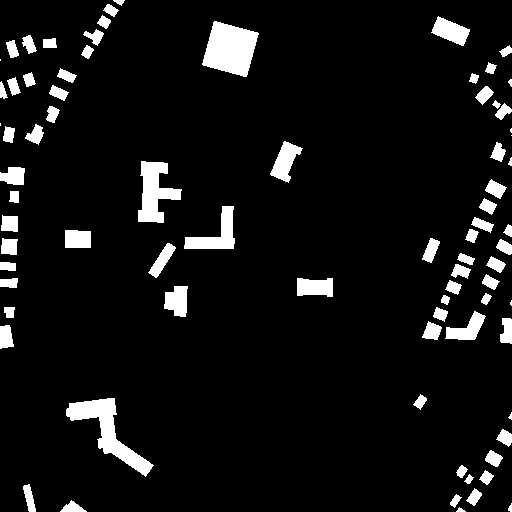} & &
        \includegraphics[width=\linewidth]{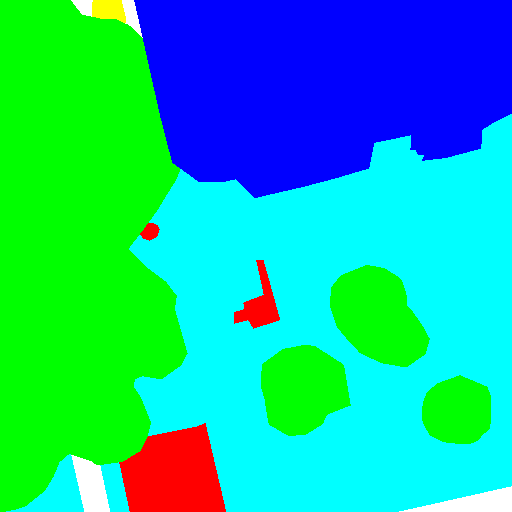} & \includegraphics[width=\linewidth]{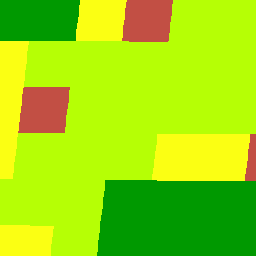} & \includegraphics[width=\linewidth]{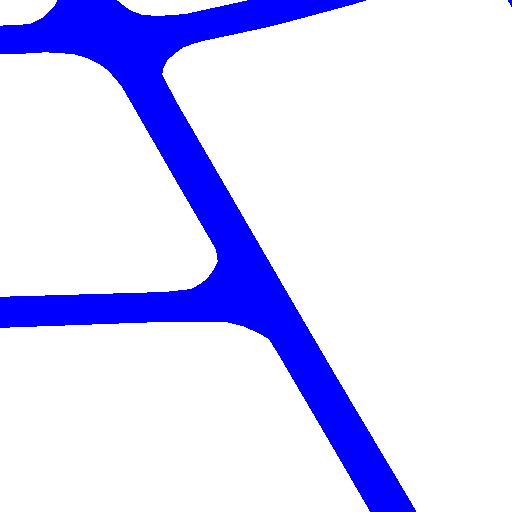} & \includegraphics[width=\linewidth]{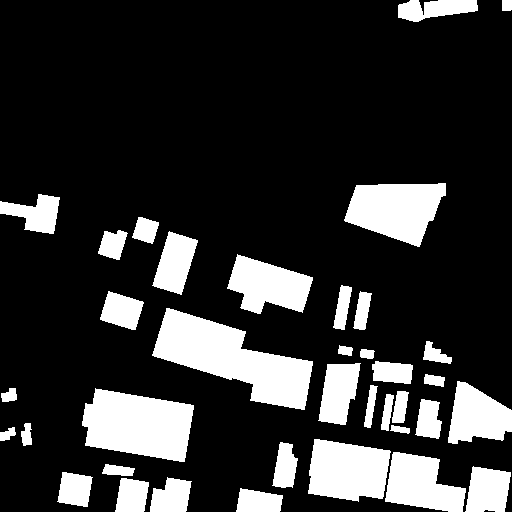} & \\

        \vspace{-3pt}(3) & \includegraphics[width=\linewidth]{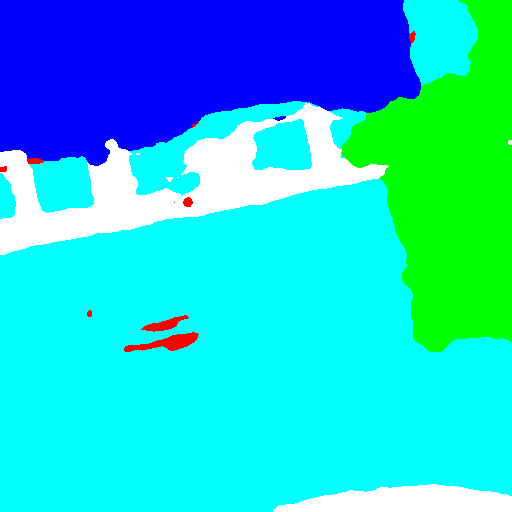} & \includegraphics[width=\linewidth]{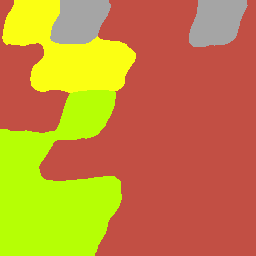} & \includegraphics[width=\linewidth]{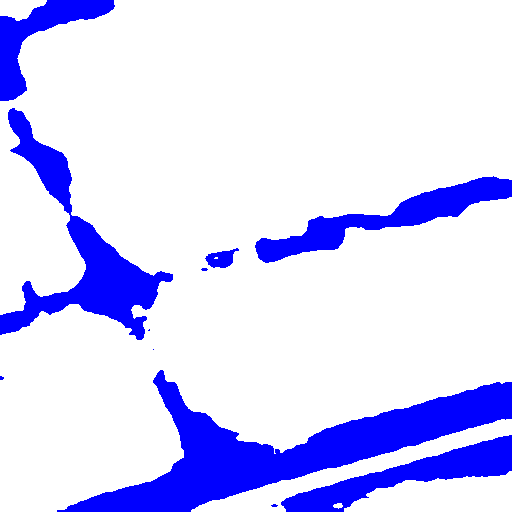} & \includegraphics[width=\linewidth]{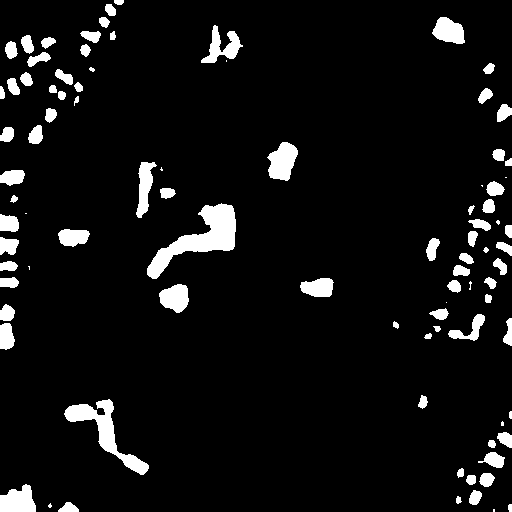} & &
        \includegraphics[width=\linewidth]{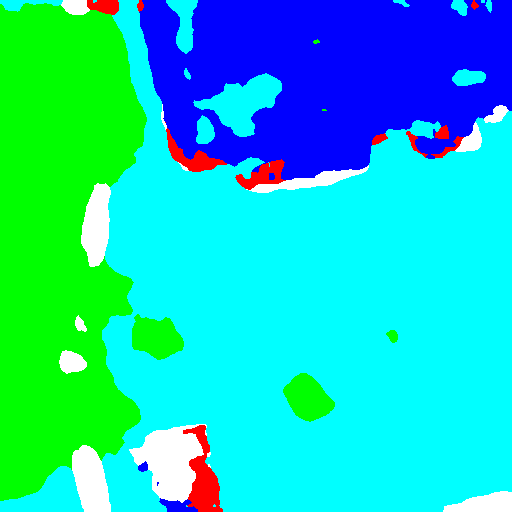} & \includegraphics[width=\linewidth]{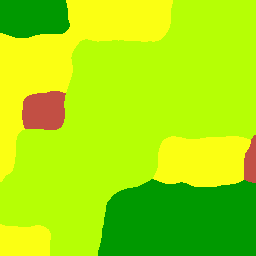} & \includegraphics[width=\linewidth]{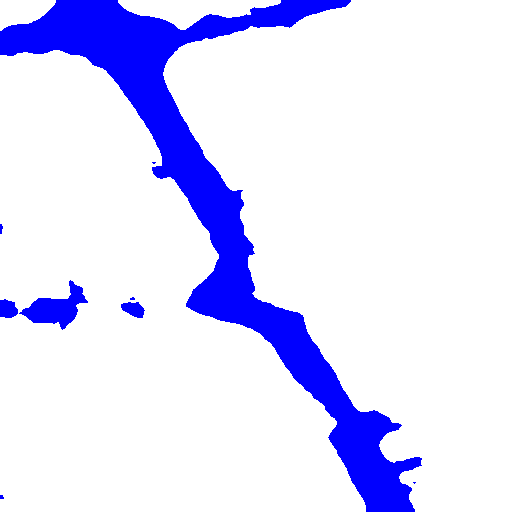} & \includegraphics[width=\linewidth]{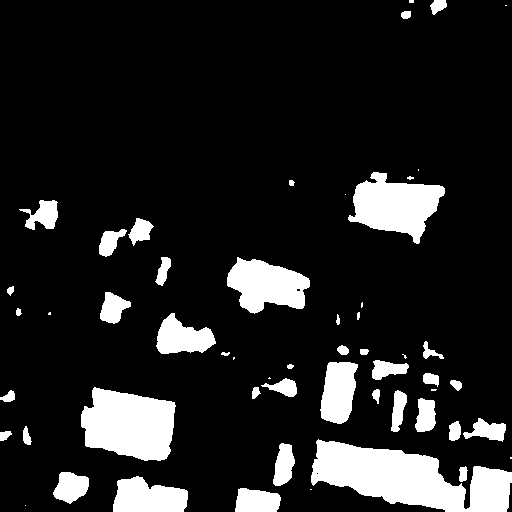} & \\
        & \centering 64.08\% & \centering	80.99\% & \centering	81.40\% & \centering	76.12\% & & \centering	41.88\% & \centering	70.08\% & \centering	85.33\% & \centering	75.90\% & \\

        \vspace{-3pt}(4) & \includegraphics[width=\linewidth]{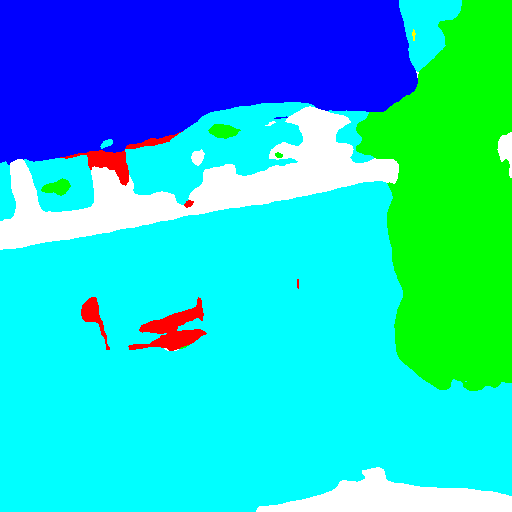} & \includegraphics[width=\linewidth]{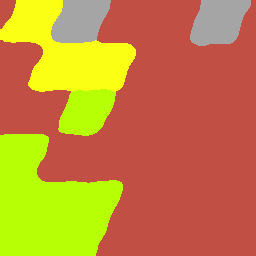} & \includegraphics[width=\linewidth]{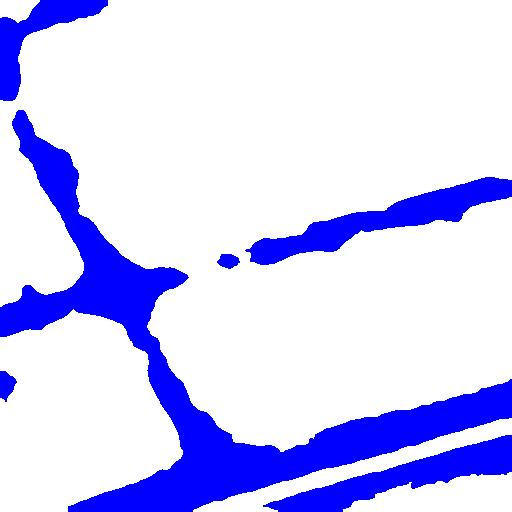} & \includegraphics[width=\linewidth]{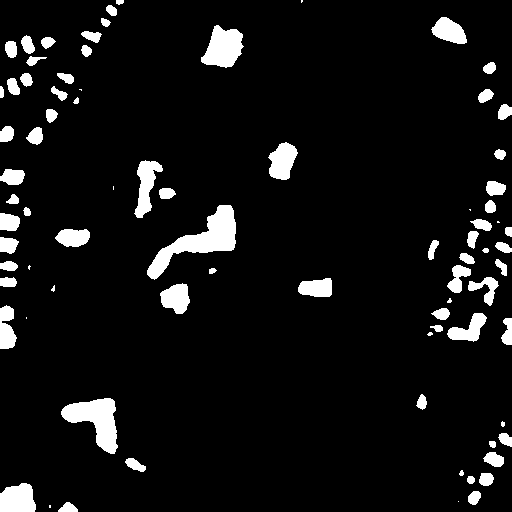} & &
        \includegraphics[width=\linewidth]{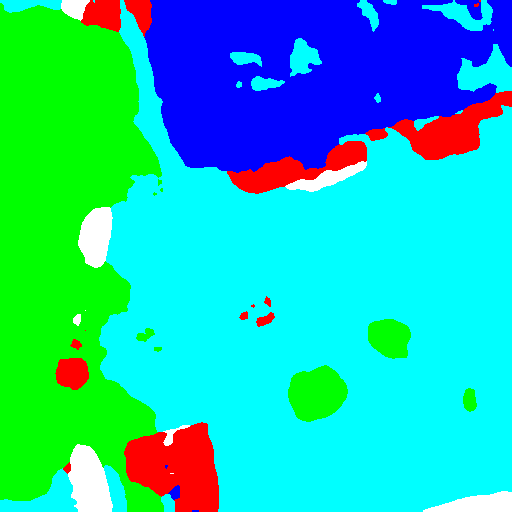} & \includegraphics[width=\linewidth]{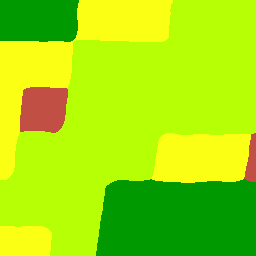} & \includegraphics[width=\linewidth]{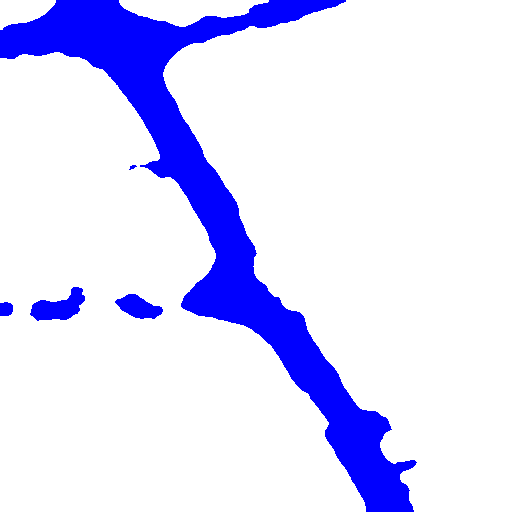} & \includegraphics[width=\linewidth]{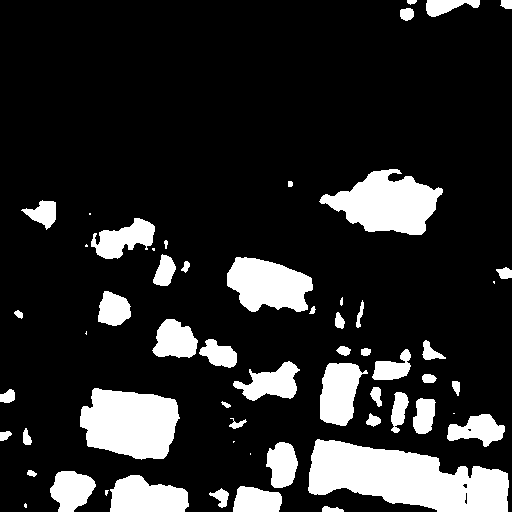} & \\

        & \centering 69.24\% & \centering	85.86\% & \centering	85.44\% & \centering	78.03\% & & \centering	47.18\% & \centering	78.21\% & \centering	86.93\% & \centering	77.44\% &  \\

        \vspace{-3pt}(5) & \includegraphics[width=\linewidth]{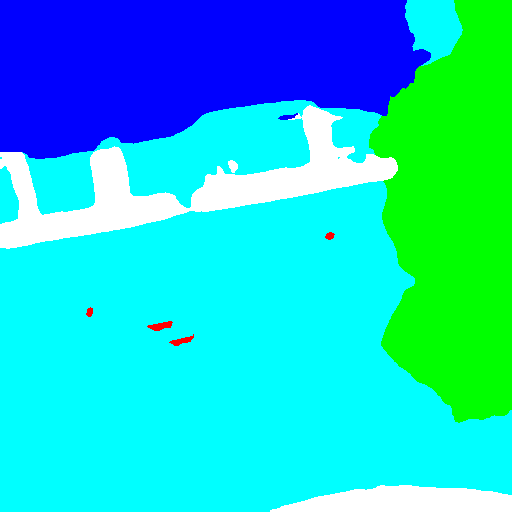} & \includegraphics[width=\linewidth]{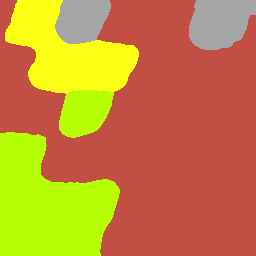} & \includegraphics[width=\linewidth]{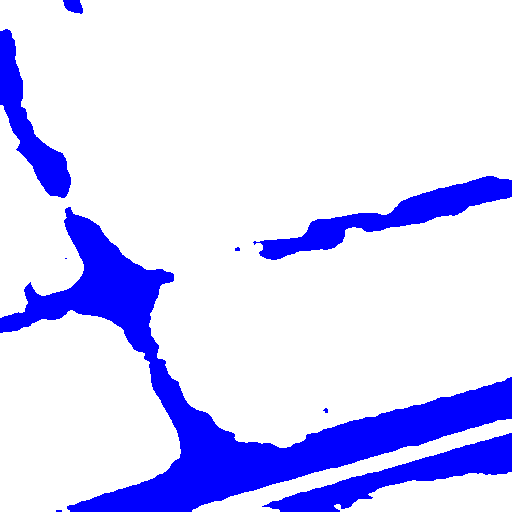} & \includegraphics[width=\linewidth]{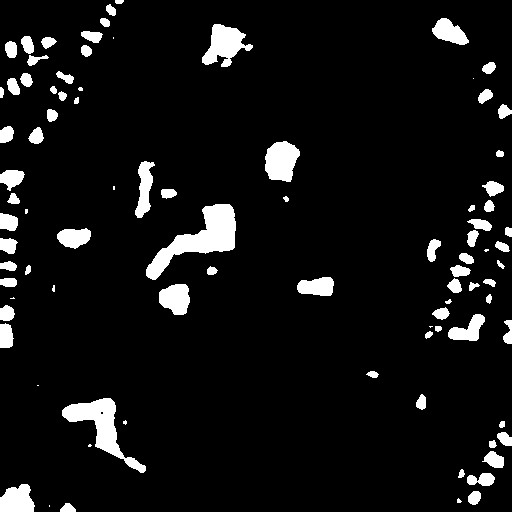} & &
        \includegraphics[width=\linewidth]{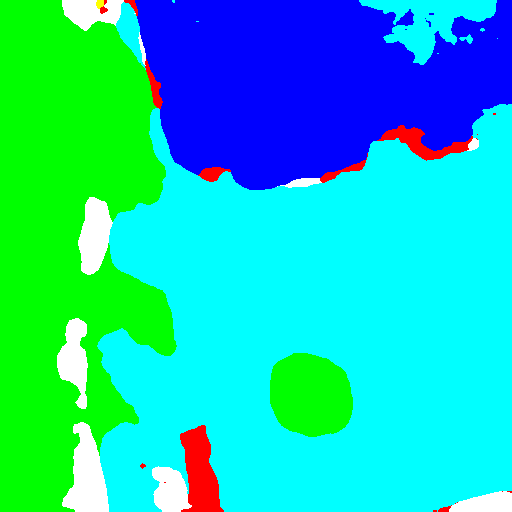} & \includegraphics[width=\linewidth]{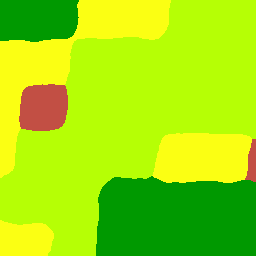} & \includegraphics[width=\linewidth]{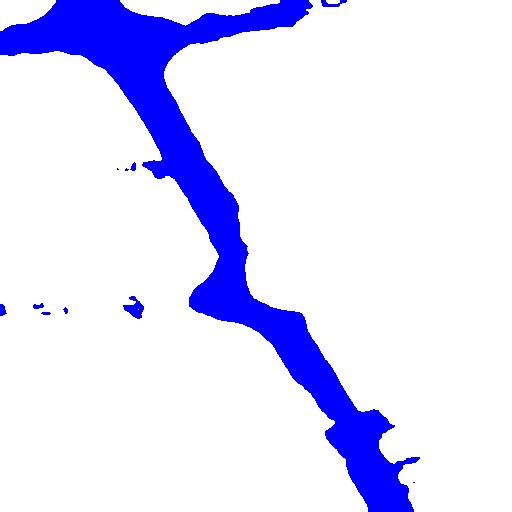} & \includegraphics[width=\linewidth]{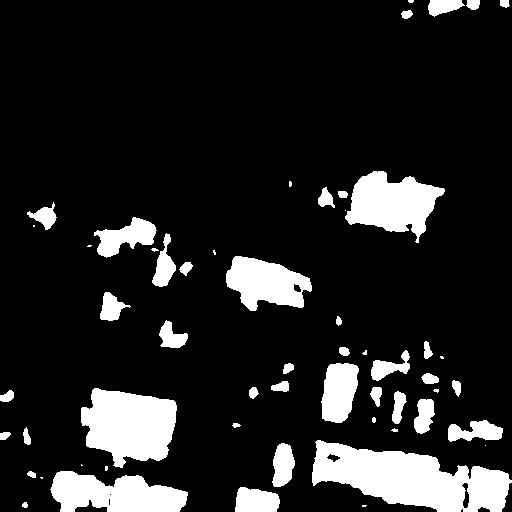} & \\

         & \centering 68.03\% & \centering	82.38\% & \centering	81.57\% & \centering	76.56\% & & \centering	47.41\% & \centering	76.66\% & \centering	83.61\% & \centering	72.72\% & \\

        \vspace{-3pt}(6) & \includegraphics[width=\linewidth]{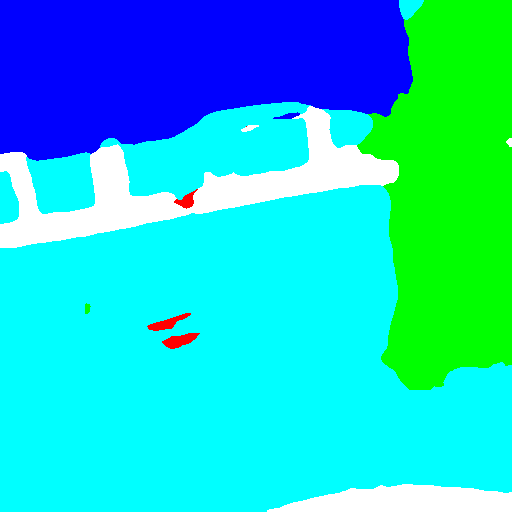} & \includegraphics[width=\linewidth]{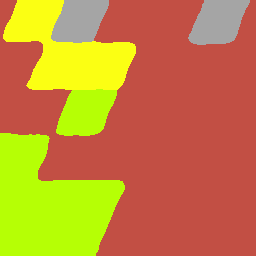} & \includegraphics[width=\linewidth]{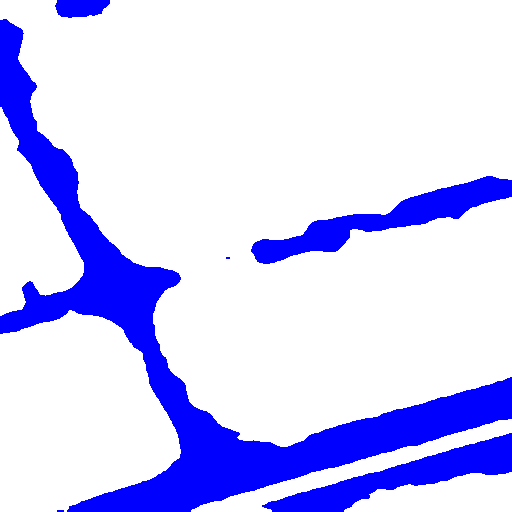} & \includegraphics[width=\linewidth]{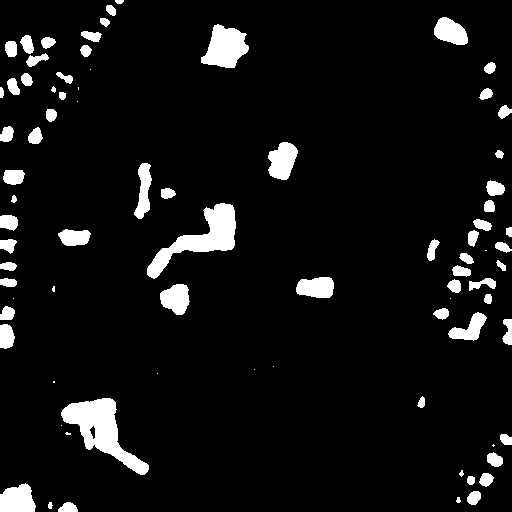} & &
        \includegraphics[width=\linewidth]{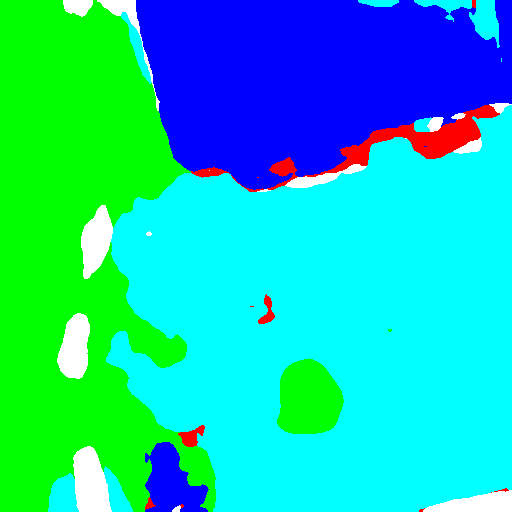} & \includegraphics[width=\linewidth]{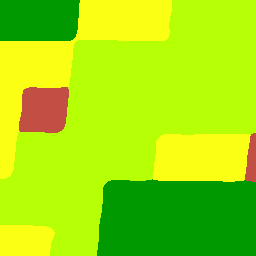} & \includegraphics[width=\linewidth]{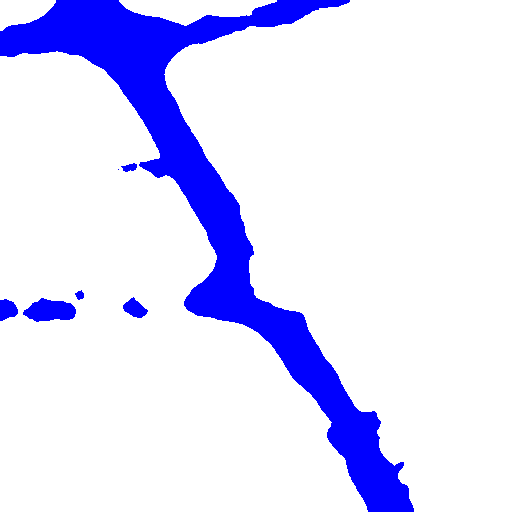} & \includegraphics[width=\linewidth]{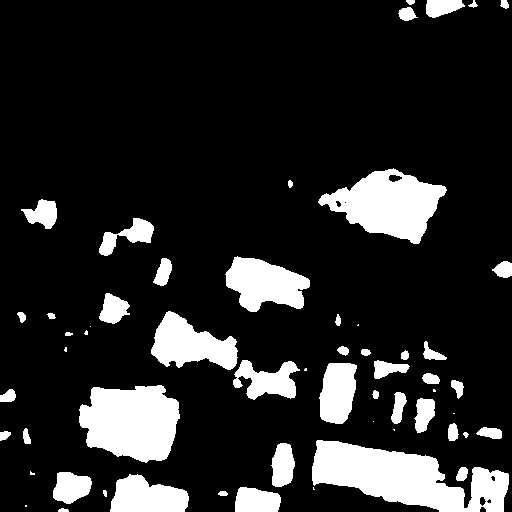} & \\

        & \centering 67.19\% & \centering	87.44\% & \centering	85.14\% & \centering	77.48\% & & \centering	45.06\% & \centering	79.39\% & \centering	86.21\% & \centering	77.09\% & \\

        \vspace{-3pt}(7) & \includegraphics[width=\linewidth]{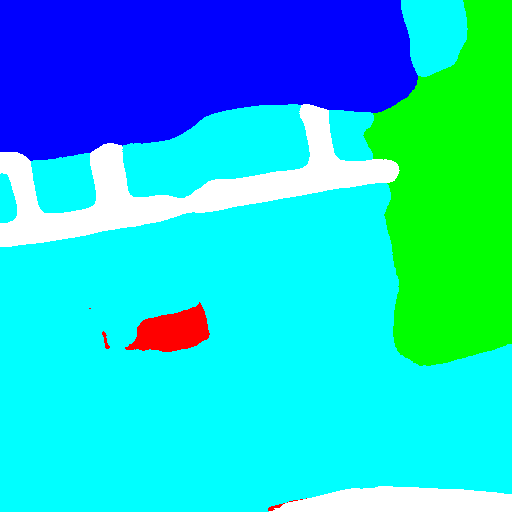} & \includegraphics[width=\linewidth]{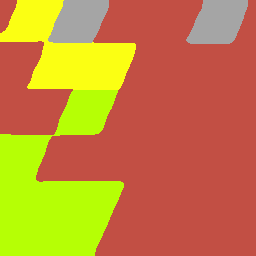} & \includegraphics[width=\linewidth]{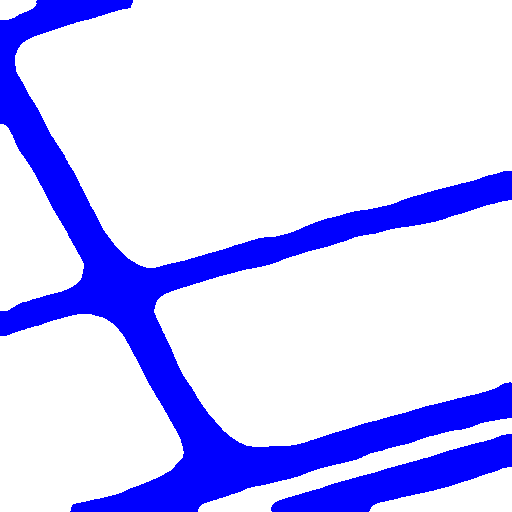} & \includegraphics[width=\linewidth]{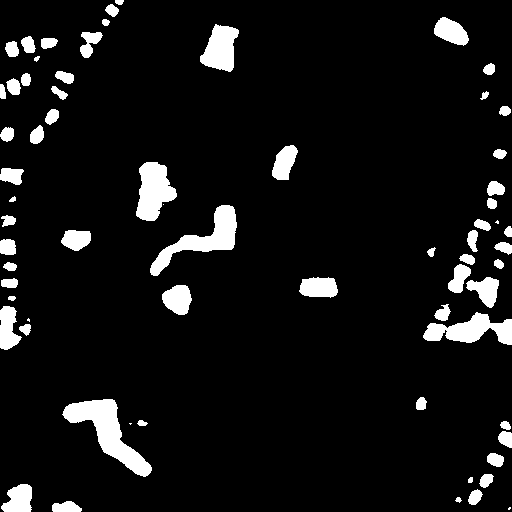} & &
        \includegraphics[width=\linewidth]{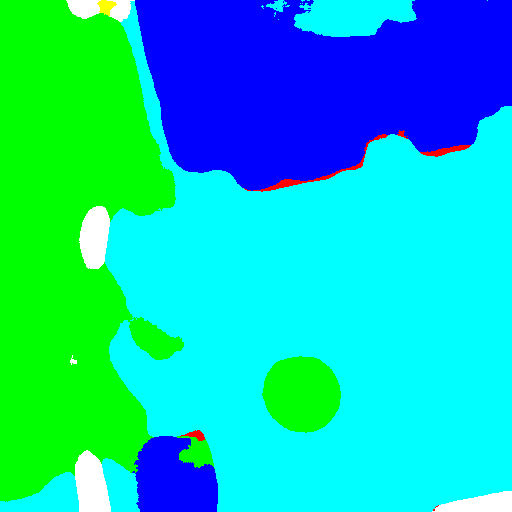} & \includegraphics[width=\linewidth]{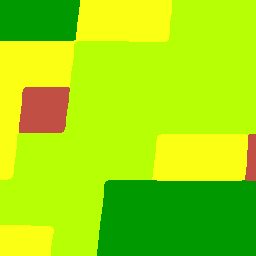} & \includegraphics[width=\linewidth]{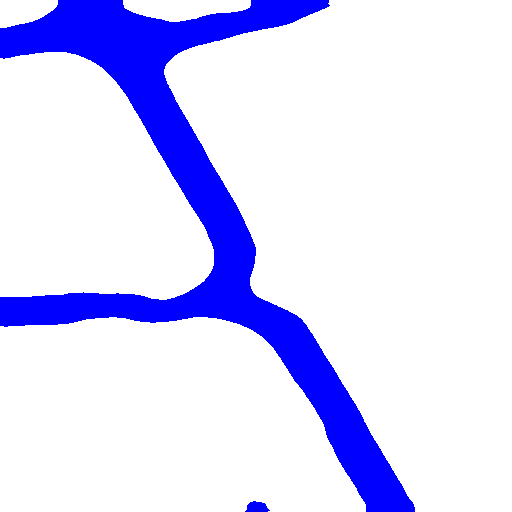} & \includegraphics[width=\linewidth]{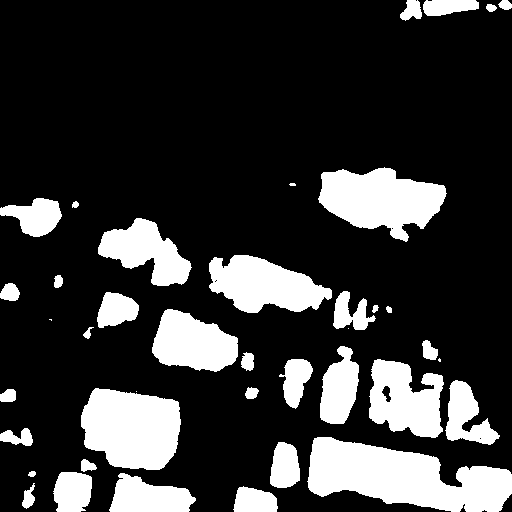} & \\
        & \centering 74.81\% & \centering	88.06\% & \centering	94.67\% & \centering	79.10\% & & \centering	51.82\% & \centering	80.28\% & \centering	92.95\% & \centering	78.71\% & \\

        \vspace{-3pt} (8) & \includegraphics[width=\linewidth]{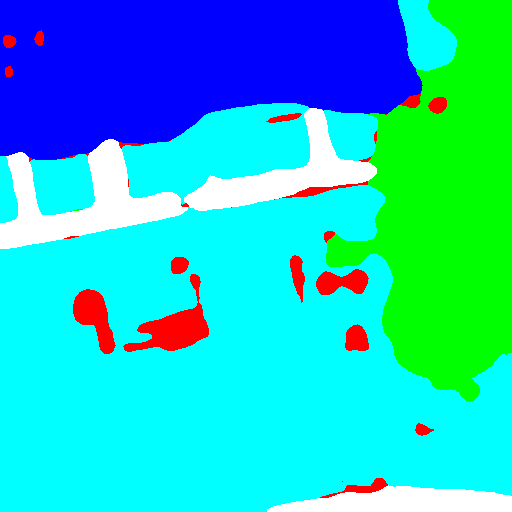} & \includegraphics[width=\linewidth]{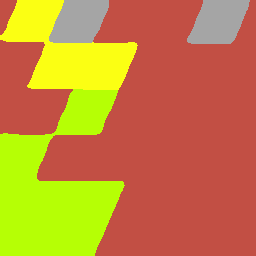} & \includegraphics[width=\linewidth]{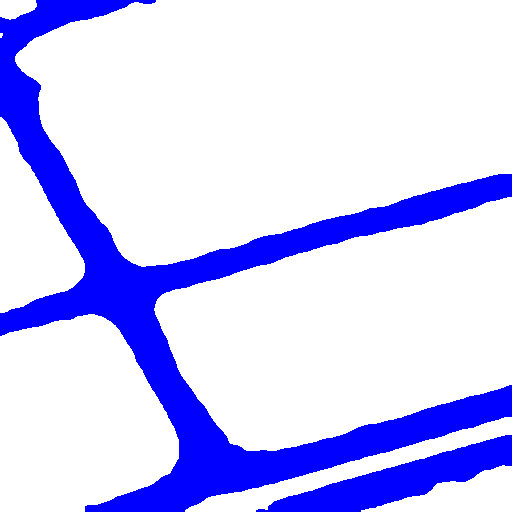} & \includegraphics[width=\linewidth]{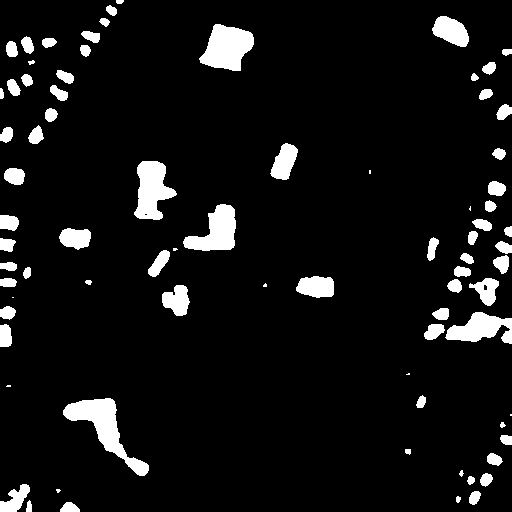} & &
        \includegraphics[width=\linewidth]{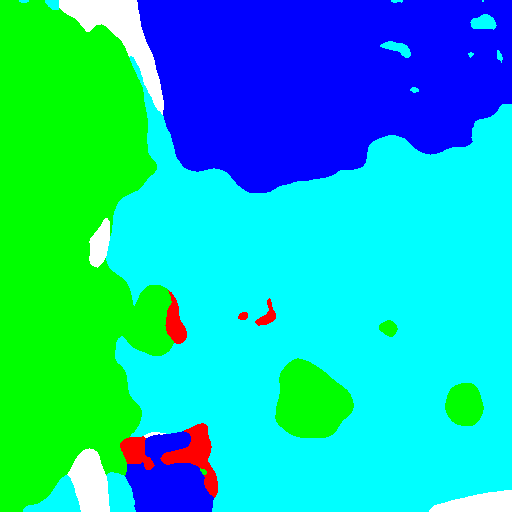} & \includegraphics[width=\linewidth]{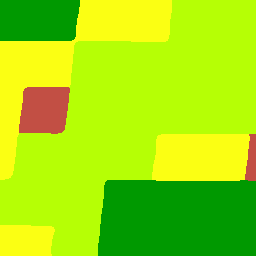} & \includegraphics[width=\linewidth]{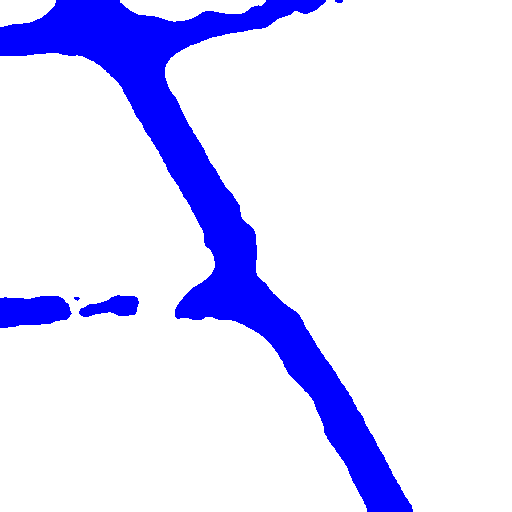} & \includegraphics[width=\linewidth]{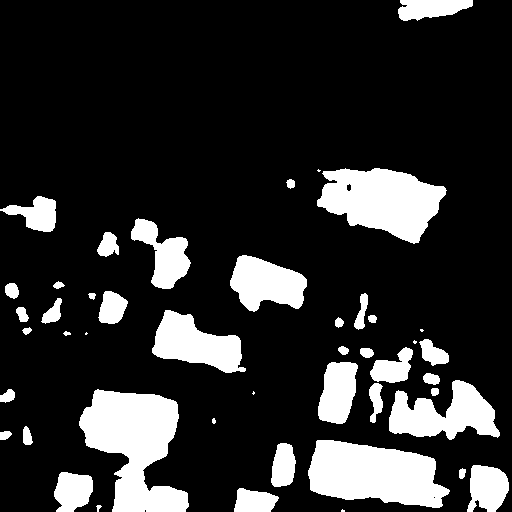} & \\

        & \centering 73.39\% & \centering	88.32\% & \centering	94.13\% & \centering	81.63\% & & \centering	50.43\% & \centering	80.12\% & \centering	90.83\% & \centering	78.89\% &\\

        \vspace{-3pt}(9) & \includegraphics[width=\linewidth]{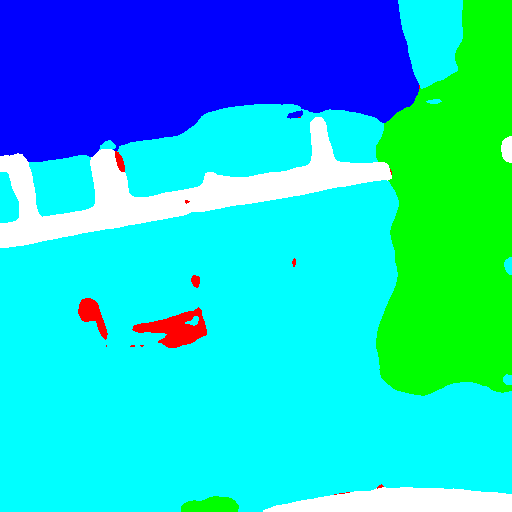} & \includegraphics[width=\linewidth]{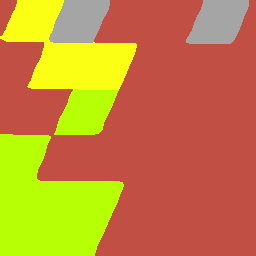} & \includegraphics[width=\linewidth]{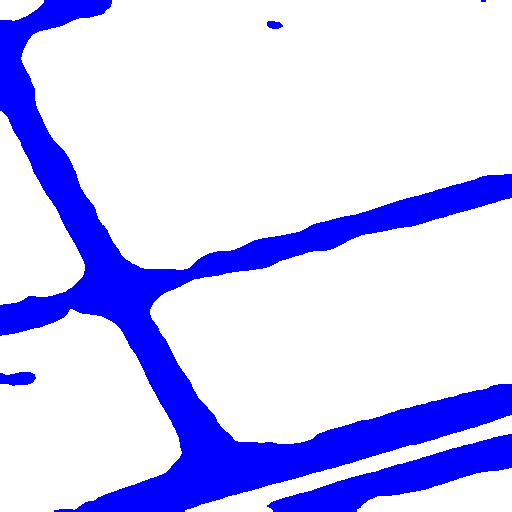} & \includegraphics[width=\linewidth]{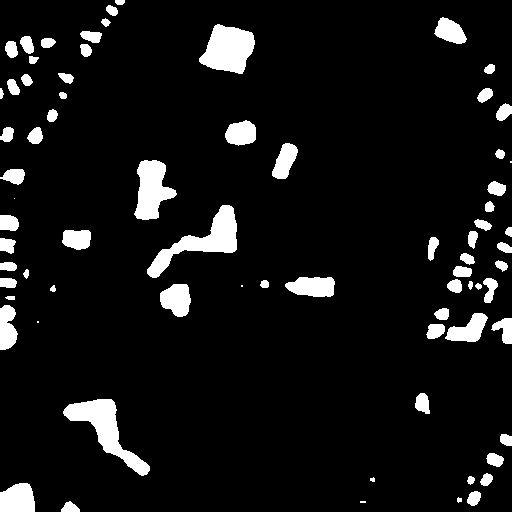} & &
        \includegraphics[width=\linewidth]{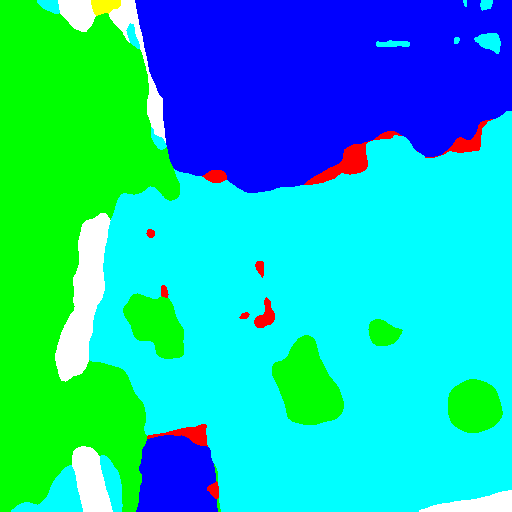} & \includegraphics[width=\linewidth]{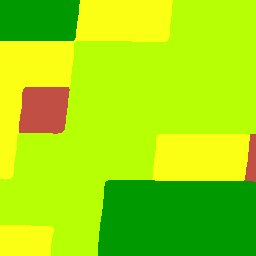} & \includegraphics[width=\linewidth]{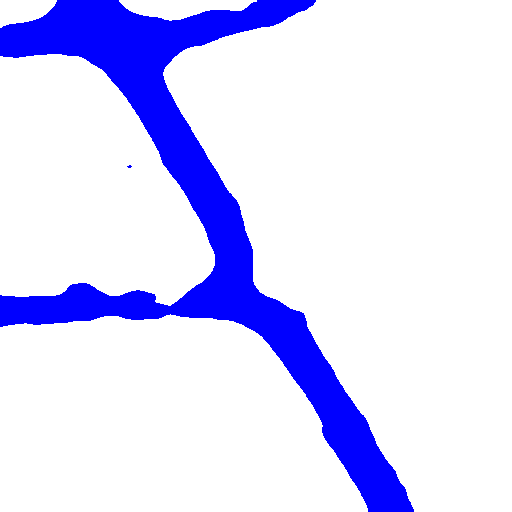} & \includegraphics[width=\linewidth]{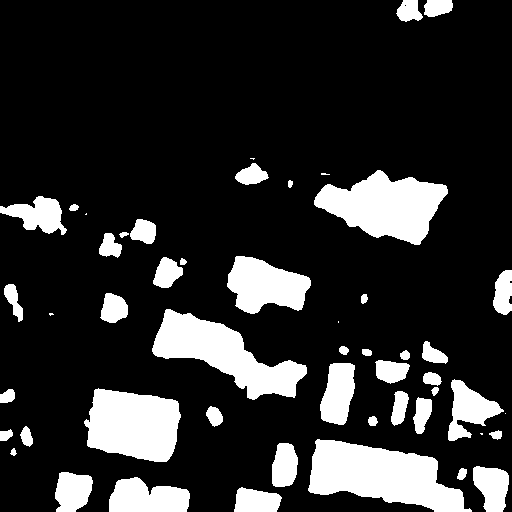} & \\
        & \centering 76.13\% & \centering	88.30\% & \centering	91.79\% & \centering	80.34\% & & \centering	55.21\% & \centering	80.31\% & \centering	92.04 & \centering	80.87\% & \\ 

    \end{tabular} \vspace{-6pt} 
     \caption{Visual results of each method on four datasets. The mIoU score of the prediction for each method is shown in its sub-caption. The numbers in the first column are -- (1): Image (2): Ground Truth (3): MT (4): CCT (5): GCT (6): CPS (7): DiverseModel (8): DiverseHead(DT) (9): DiverseHead(DF)}
    \label{fig:Visual_results2}
\end{figure*}

The visible results are presented in Figures \ref{fig:Visual_results2}. For each dataset, we randomly selected 2 cases of the original image, its ground truth, and predictions for the proposed methods and various classic methods. Also, the mIoU scores are calculated for all predictions with the ground truth for each case, which are shown in the sub-caption of each prediction. Based on the IoU values of visual predictions, the highest score is obtained by either DiverseModel or DiverseHead (DT\&DF) for most cases. Especially for cases of the RoadNet dataset, both the DiverseModel and DiverseHead families achieve a mIoU of over 90\% surpassing other methods by at least 6.35\%. Visually, the segmentation maps of DiverseModel and DiverseHead display better overall similarity to the ground truth than other methods.

\section{Conclusion}\label{sec:conc}
In this paper, we proposed a lightweight and efficient semi-supervised learning approach based on a multi-head structure called DiverseHead. Based on the multi-head structure, we provide two perturbation methods, namely dynamic freezing and dropout. Taking inspiration from the theory of bagging, a voting mechanism is proposed to generate beneficial pseudo-labels in the training stage. This simple and lightweight semi-supervised learning framework shows competitive performance for the segmentation of remote sensing imagery. Also, It can be easily combined with state-of-the-art SSL methods and further improve their performance. Furthermore, we conducted further analysis and additional evaluation on the previously proposed multi-network-based semi-supervised learning method known as DiverseModel. Based on the results obtained from the aforementioned four remote sensing datasets, DiverseHead and DiverseModel demonstrate comparable performance while significantly outperforming various classic semi-supervised learning frameworks. From the application perspective, the proposed DiverseNet could theoretically be utilised for a wide range of image-based tasks, including medical imaging, remote sensing, and natural image segmentation. Having evaluated the methods on datasets with multi-band images, our approach is likely to perform competitively in MRI and other multi-band image segmentation tasks. We will explore more applications using the proposed method in the future. For technical improvement, future work could explore investigating the integration of the two proposed and other perturbation strategies in SSL, and designing more sophisticated interaction mechanisms among component models and optimising the hyperparameters to enhance their synergy.

\bibliographystyle{IEEEtran}
\bibliography{ref}

\end{document}